\documentclass{article} 
\usepackage{iclr2027_conference}
\usepackage{times}

\usepackage{hyperref}
\usepackage{url}
\usepackage{graphicx}
\usepackage{wrapfig}
\usepackage{xcolor}
\usepackage{xspace}
\usepackage{enumitem}
\usepackage{array}
\iclrfinalcopy

\title{Before the Rollout Ends: Early Terminal Reward Prediction for Long-horizon Coding Agents}

\author{    
    Jihan Yao\textsuperscript{1}, 
    Sihan Zeng\textsuperscript{2}, 
    Shangbin Feng\textsuperscript{1},
    Zhiyuan Fan\textsuperscript{3}, 
    Banghua Zhu\textsuperscript{1}, 
    Yulia Tsvetkov\textsuperscript{1} \\
    \textsuperscript{1}University of Washington  \ \ \ 
    \textsuperscript{2}Independent Researcher   \ \ \ 
    \textsuperscript{3}HKUST   \ \ \ \\
    \texttt{jihany2@cs.washington.edu}
}

\usepackage{booktabs}
\usepackage{multirow}
\usepackage{amsmath,amssymb}
\usepackage{algorithm}
\usepackage{algpseudocode}
\usepackage[most]{tcolorbox}

\newcommand{\todo}[1]{{\color{red}\textbf{[TODO:]}}\xspace}

\definecolor{promptcardframe}{HTML}{C8CED8}
\definecolor{promptcardfill}{HTML}{F5F7FA}
\newtcblisting{promptcard}[1]{enhanced,breakable,listing only,listing engine=listings,title={#1},colback=promptcardfill,colframe=promptcardframe,coltitle=black,fonttitle=\bfseries,boxrule=0.5pt,arc=1mm,left=1mm,right=1mm,top=1mm,bottom=1mm,listing options={basicstyle=\ttfamily\scriptsize,breaklines=true,breakatwhitespace=false,breakindent=0pt,breakautoindent=false,columns=fullflexible,keepspaces=true,showstringspaces=false}}

\begin{document}

\maketitle

\begin{abstract}
Long-horizon coding agents receive verifiable rewards only after completing expensive sequences of tool calls. This increases inference cost, amplifies early wrong hypotheses, and can lead to sparse terminal reward and unstable training. We introduce Contextual Early Reward (CER), which predicts terminal reward through behavioral evidence in a trajectory prefix. CER synthesizes adaptive rubrics specific to the current task and stage through experiences summarized from related historical tasks. In test-time scaling on SWE-bench Verified, CER improves RM@8 over the strongest baseline by 4.2 percentage points (pp) on Nemotron 3 Ultra and 2.0 pp on Qwen 3.6 27B; on Nemotron, it takes only 15.3\% tokens to match the best baseline performance. In RL training experiments, CER exceeds full-rollout TMax by 1.9 pp while using 52.7\% fewer online policy-and-judge tokens. Together, CER provides an interpretable, efficient, and dense evaluation method for long-horizon coding agents.
\end{abstract}

\section{Introduction}
\label{sec:introduction}

As LLM coding agents become ubiquitous and take on increasingly realistic tasks, their interaction horizons continue to grow. Longer horizons expand what agents can accomplish, but they also significantly delay evidence about whether a trajectory is successful. This increases both inference and learning costs. Prior work finds that agents stop only after many unnecessary interactions and that iterative reasoning and failure retries consume substantial compute \citep{luo2026agentic}, while predictive termination reduces such waste of resources \citep{pham2026agentstop}. Moreover, autoregressive agents are conditioned by their own previous decisions: an incorrect hypothesis changes which files are inspected and which edits appear plausible. Later actions may reinforce the same hypothesis, turning an early mistake into a long, coherent, but unsuccessful trajectory \citep{zhang2026agentforesight}. Delayed feedback also weakens post-training. A terminal binary reward can assign the same zero to two very different trajectories: one may reproduce the bug and localize the responsible module but fail to finish, while the other may browse unrelated files or edits without diagnosis. Encouragingly, previous work finds that intermediate progress reward can provide credit that sparse outcomes omit and is beneficial for overall training \citep{choudhury2025process}.

We therefore study \emph{early terminal-reward prediction}. Given an unfinished prefix, the objective is to estimate its future terminal reward without completing the trajectory. The forecaster can use the observed tool outputs, and current workspace diff, but not future actions, the final patch or evaluation. This differs from a process reward model, which evaluates the reward of a local action. We leverage what the interaction so far implies about the trajectory's eventual success.

We propose \textbf{Contextual Early Reward (CER)}. The intuition is that early actions are informative: reproducing a failure, validating an edit, or recovering from a bad hypothesis can strongly indicate future success, whereas unsupported edits, repeated browsing, and no self-testing can indicate failure. However, the value of each behavior depends on the task and its stage, or is ``contextual'' in other words. CER leverages the semantic differences within a group of candidate prefixes, constructs one shared weighted rubric set, and scores every prefix under the same criteria while requiring criterion-level evidence from the visible interaction. CER instantiates this interface in two regimes: without verifiable signals, it retrieves judging experience from related historical tasks and synthesizes adaptive rubrics for the current task and stage; when verifiable signals are available, it distills successful and unsuccessful patterns into stronger task-specific offline rubrics that serve for online training.

We use early rewards in two downstream tasks on SWE-bench Verified \citep{jimenez2024swe}. For test-time scaling (TTS), the agent generates candidate continuations from unfinished trajectories with a limited budget. CER ranks the visible continuations and directs subsequent generation toward the more promising candidates, concentrating the inference budget without first completing every alternative. For reinforcement learning (RL) training, we sample groups of unfinished trajectories. CER scores each prefix, transforms the scores into group-relative reward, and optimizes the policy without training on complete trajectories or querying the terminal verifier online; the task-specific rubrics are constructed offline from historical rollouts, so CER provides a verifier-free online reward.

In the TTS experiment on SWE-bench Verified, CER raises RM@8 over the strongest baselines from 63.4\% to 67.6\% on Nemotron 3 Ultra \citep{blakeman2026nemotron} and from 69.4\% to 71.4\% on Qwen 3.6 27B \citep{qwen2026qwen36}. With only 15.3\% and 74.2\% of the required tokens on \textsc{Nemotron 3 Ultra} and \textsc{Qwen 3.6 27B}, CER can match the best baseline performance. In a three-fold RL experiment on Qwen 3.5 9B, CER exceeds complete-rollout TMax baseline by a small margin (51.6\% versus 49.7\%) while using 52.7\% fewer online tokens during RL. In further analysis, we find that experience transfer is directional: experience summarized from stronger models can benefit weaker models, with Qwen 3.6 27B-derived experience improving Nemotron 3 Ultra by 1--2 points across each budget cutoff. We also find only roughly 60--65\% pairwise agreement between rubric rewards and terminal rewards in our RL experiments, indicating that the two signals are not interchangeable; instead, their downstream outcomes are comparable and the signals may be combined. Together, these findings show that CER can support efficient policy training and effective rubric-based evaluation of coding agents as a supplement to test-based verification.

\begin{figure*}[t]
  \centering
  \includegraphics[width=0.99\textwidth]{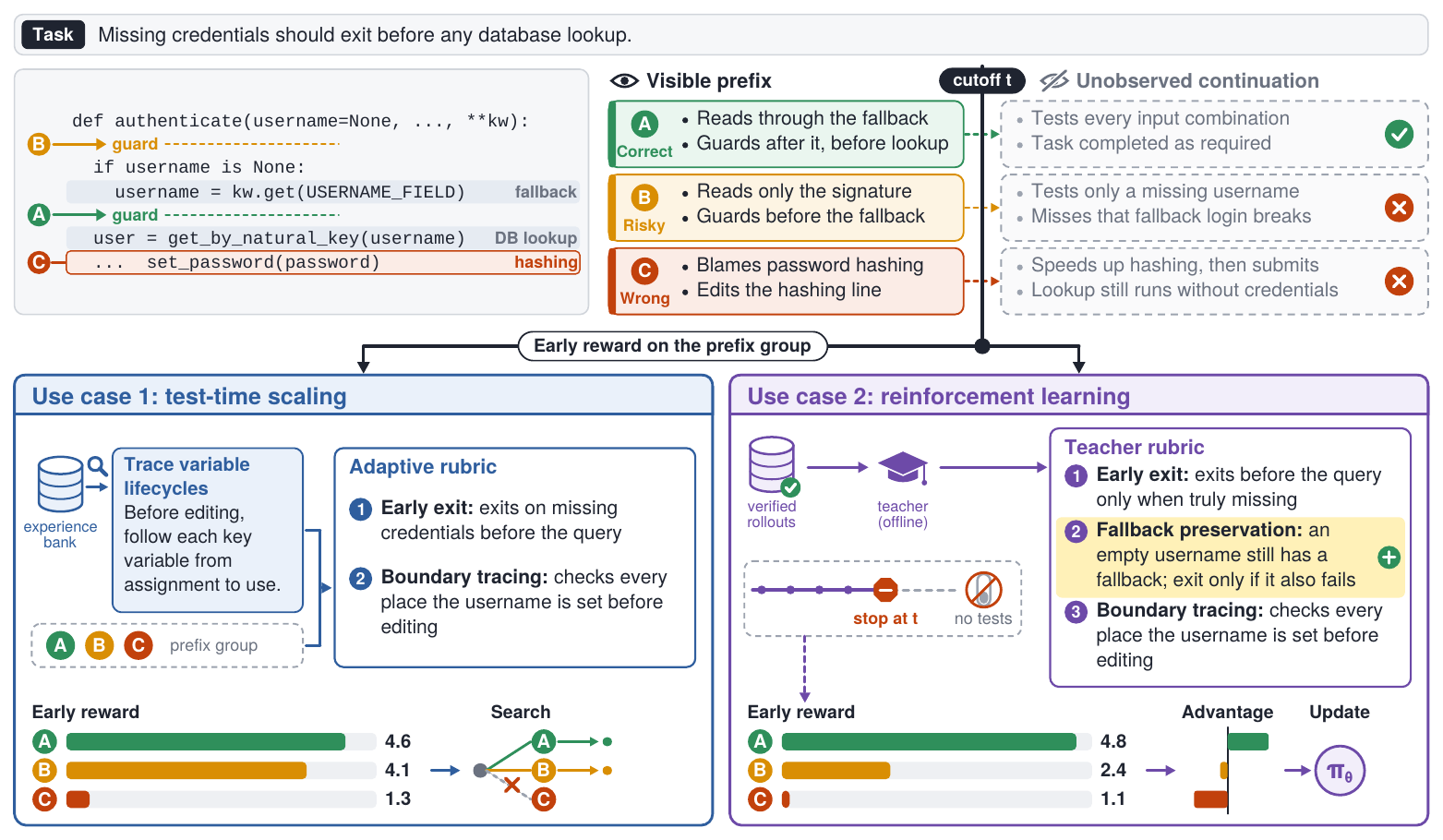}
  \caption{\textbf{CER overview.} At cutoff $t$, CER scores a group of unfinished rollout prefixes with contextual rubrics to predict their final rewards. For test-time scaling, rubrics are generated from retrieved cross-task experience and comparable prefix samples, and only top-ranked prefixes are expanded. For reinforcement learning, online rollouts stop at $t$, and teacher-summarized rubrics score each prefix to provide group-relative advantages for the policy update. With access to verified rollouts, teacher rubrics provide finer-grained and more accurate signals.}
  \label{fig:method}
  \vspace{-20pt}
\end{figure*}

\section{Related Work}
\paragraph{Process reward for coding agents.} Previous work uses process reward models (PRMs) to provide dense feedback on intermediate behavior as a supplement to test-based verifiers. AgentPRM estimates step values from Monte Carlo continuations \citep{choudhury2025process}. OpenHands Critic trains a critic model on external SWE trajectories with temporal-difference targets derived from terminal unit-test outcomes \citep{openhands2025critic}. Other work assumes internal state access of agent LLMs, extracting prefix-aware rewards from a policy model's internal representations \citep{zhou2026correctness, kim2026pair}. Monte Carlo PRMs estimate state values but require impractical sampled suffix rollouts for long-horizon agents, while learned prefix critics can be sensitive to policy and scaffold shifts, as shown in our later experiments. CER instead constructs task- and stage-conditioned evaluation criteria without training a separate value model or probe that requires white-box access.

\paragraph{Execution-free verification for coding agents.} Test execution remains an important efficiency bottleneck for agentic training. Moreover, recent studies identify false positives and negatives in SWE-style evaluation \citep{wang2025solved, yu2025utboost} and a gap between verifier rewards and user intent \citep{wang2026verification}, indicating that test verifiers are insufficient rewards. One solution is execution-free reward models, which score software-agent behavior without running a hidden test suite. SWE-RM trains a generative resolved/unresolved classifier from large mixtures of strong LLM trajectories \citep{shum2026swe}. SIMIA replaces real environment implementations with strong external proprietary model as simulators that generate tool feedback and rewards for agent training \citep{li2025simulating}. Other methods leverage rubrics to evaluate agent trajectories. OpenHands Rubrics learns a fixed set of behavioral features from real-world human feedback \citep{wang2026rubric}. Agentic Rubrics \citep{raghavendra2026agentic} asks an expert agent to inspect a repository and construct a context-grounded checklist for evaluating candidate trajectories. SWE-PRM \citep{gandhi2025agents} inspects a single trajectory with strong external proprietary model to identify early deficiency. AMARIS retrieves diagnostics from a persistent training history to update rubrics but has not been tested on agentic tasks \citep{wu2026amaris}. However, these methods either require complete trajectories and cannot be used for early prediction, or fail to explore the rich judging context and memory.

\section{Methodology}
\subsection{CER}

Let $x$ denote a problem statement and $h_t=(x,a_1,o_1,\ldots,a_t,o_t)$ an agent trajectory observed after $t$ interaction turns. Conventional verification assigns a terminal reward $y(\tau_T)\in\{0,1\}$ only after the agent submits a patch and the evaluator executes its tests. At turn $t<T$, the ideal value for current state under policy $\pi$ is $V_{x, \pi}(h_t)=\Pr_{\tau_T\sim\pi(\cdot\mid h_t)}[y(\tau_T)=1]$.

One cannot learn a value model without complete trajectories and terminal rewards. We instead construct a task-conditioned rubric set $\mathcal R_{x, t}$ and use a judge model $J$ to predict an early reward $s_x(h_t)$ from the visible prefix alone. $\mathcal R_{x, t}=\{(r_{x, t}^{k},w_{x, t}^{k})\}_{k=1}^{K}$ is a set of weighted, task-relevant criteria. Each $r_{x, t}^{k}$ describes an observable property of a good trajectory, such as whether the agent localized the relevant code, checked regressions after editing, or recognized and recovered from an unproductive direction. We refer to this as \textbf{Contextual Early Reward (CER)}. Given an unfinished prefix, the judge returns both a criterion score $r_{x, t}^{k}(h_t)$ and supporting evidence from the visible interaction: 
\begin{equation}
 s_x(h_t)=\frac{\sum_{k=1}^{K} w_{x, t}^{k} \ r_{x, t}^{k}(h_t)}
                 {\sum_{k=1}^{K} w_{x, t}^{k}}.
 \label{eq:rubric-score}
\end{equation}
Ideally, $s_x(h_t)=V_{x, \pi}(h_t)$, under which $s_x$ is a perfectly calibrated predictor of terminal reward. However, this requirement is unnecessarily strict for an LLM judge that is not trained to estimate continuation probabilities directly. Considering that the judge model compares candidates within a group $H_t=\{h_t^{1},\ldots,h_t^{m}\}$ sampled for the same task, only $s_x(h_t)=V_{x, \pi}(h_t)+b_{x,t}$ is required for some group-specific $b_{x,t}$ (see Appendix \ref{app:advantage-bound} for proof). This means $s_x(h_t)$ needs to preserve only the pairwise relative differences. Group normalization removes the offset terms, so CER need not be calibrated across tasks or stages. For downstream tasks that depend only on the relative order within the group, this condition can be relaxed further. Thus it is sufficient to have
\begin{equation}
 \operatorname{sgn}(s_x^i(h_t)-s_x^j(h_t))=\operatorname{sgn}(V_{x, \pi}^i(h_t)-V_{x, \pi}^j(h_t)),\qquad \forall i,j\ \text{with}\ i\neq j.
 \label{eq:ranking}
\end{equation}
This means that when $s_x(h_t)$ preserves the judgment order of each prefix pair within the group, CER selects the same top candidates as the true terminal values even if their numerical gaps are distorted. TTS only requires this relaxation, since it uses the scores solely to rank candidates.

\subsection{CER for Test-Time Scaling}
\label{sec:experience-rubrics}
To enable reliable early terminal-reward prediction, CER takes advantage of two sources of enriched judging context. First, continuations sampled from a common prefix form a controlled comparison group. Their behavioral differences provide local reference that improves within-group calibration. Second, CER retrieves experience $\mathcal E_x$ from historical tasks or agent stages similar to the current one. Because these experiences are summarized and corrected against verifiable terminal rewards, they provide empirically grounded judging principles and reduce the judge model's intrinsic bias. 

For each historical task $x$ and search round $r$, we form group dataset $G_{x,r}=\{(p_r,h_i,\hat{s}_i,s_i)\}_{i=1}^{m}$ containing the parent prefix $p_r$, trajectory continuations $h_i$, rubric-judge scores $\hat{s}_i$, and terminal verifier rewards $s_i$. We synthesize experience from every source group satisfying $\operatorname{Var}(s_1,\ldots,s_m)>0$ and pairwise judging accuracy $\operatorname{Acc}_{\mathrm{pair}}(\hat{\mathbf{s}},\mathbf{s})=\frac{\sum_{i<j}\mathbf{1}[s_i\neq s_j]\mathbf{1}[(\hat{s}_i-\hat{s}_j)(s_i-s_j)>0]}{\sum_{i<j}\mathbf{1}[s_i\neq s_j]}$ $< 0.5$. Each experience is a structured record containing a short description; an applicability context specifying the task type, agent stage, observable triggers, and abstention conditions; an actionable lesson explaining which evidence to reward or distrust; and reference rubrics that instantiate the lesson. A strong teacher model first generates experience by comparing continuations of various verifier rewards. Then the judge rescores $G_{x,r}$ and retain the candidate only if pairwise judging accuracy improves over rubric generation without it (i.e., $\operatorname{Acc}_{\mathrm{pair}}(\hat{\mathbf{s}}^{e},\mathbf{s})>\operatorname{Acc}_{\mathrm{pair}}(\hat{\mathbf{s}},\mathbf{s})$). We accept one additional experience-refinement step to sharpen its activation conditions and evidence boundaries. We construct the retriever by representing each accepted experience with keywords summarized by an LLM and keywords frequently shared between the trajectories and the experience.

CER retrieves experiences suitable for the current task and uses them to guide test-time scaling in Algorithm~\ref{alg:cer-tts}. Three components are found empirically useful. (1) \textbf{Regression stopping} removes continuations whose score falls below that of their parent prefix from further expansion, preventing the agent from spending further computation along an increasingly erroneous direction. (2) \textbf{Beam parents} expand multiple leading parents rather than only the best one, preserving continuation diversity and avoiding rapid diversity collapse. (3) \textbf{Tie breaking} generates additional discriminative rubrics conditioned only on tied continuations. This design follows the observation that the judge model tends initially to focus on major observable differences, leading to many tied judgment scores.

\begin{algorithm}[t]
\caption{CER for test-time scaling}
\label{alg:cer-tts}
\begin{algorithmic}[1]
\Require Task $x$, experience bank $\mathcal E_x$, beam width $b$, max search rounds $R$, turn budget $t$ per round
\State $P\gets[x]$, $z^\star\gets x$, $r\gets0$ \Comment{$P$ is a score-ordered stack; $z^\star$ is the current selection}
\While{$r < R$ \textbf{and} $P \neq \varnothing$}
\State Start with the first $b$ beam parents at the current depth: $\mathcal P_r\gets\operatorname{Pop}_{\min\{b,|P|\}}(P)$
\State Expand each $p\in\mathcal P_r$ to obtain $H_r=\bigcup_{p\in\mathcal P_r}\operatorname{rollout}_{m/|\mathcal P_r|}(p)$ and let $p(h)$ denote the parent of $h$
\State Retrieve relevant experience $E_r\gets\operatorname{BM-25}\ (x,\mathcal P_r,H_r;\mathcal E_x)$
\State Generate rubrics $\mathcal R_{x,r}\gets\operatorname{LLM}\ (x,\mathcal P_r,H_r,E_r)$ and let $z(h)=p(h)\oplus h$
\State Under the same $\mathcal R_{x,r}$, score every $z\in\mathcal P_r\cup\{z(h):h\in H_r\}$ by $s_{x,r}(z)=\frac{\sum_{k=1}^{K}w_{x,r}^{k}r_{x,r}^{k}(z)}{\sum_{k=1}^{K}w_{x,r}^{k}}$
\State $\mathcal T_r\gets\{z(h):\exists h'\neq h,\ s_{x,r}(z(h))=s_{x,r}(z(h'))\}$
\If{$|\mathcal T_r|>0$}
\State Generate additional tie-breaking rubrics $\mathcal R_{x,r}'\gets\operatorname{LLM}(x,\mathcal P_r,\mathcal T_r,E_r\mid\mathcal R_{x,r})$
\State Re-score tied continuations: $s_{x,r}(z(h))\gets s(z(h);\mathcal R_{x,r}'),\ \forall z(h)\in\mathcal T_r$
\EndIf
\State Keep non-regressed continuations: $C_{x,r}\gets\{z(h):h\in H_r,\ s_{x,r}(z(h))\geq s_{x,r}(p(h))\}$
\If{$C_{x,r}\neq\varnothing$} 
\State $z^\star\gets\operatorname{Top}_1(\{z(h):h\in H_r\};s_{x,r})$
\EndIf
\State $P\gets\operatorname{Push}\!\left(P,\operatorname{Top}_{\min\{2b, |C_{x,r}|\}}(C_{x,r};s_{x,r})\right)$ 
\State $r \gets r + 1$
\EndWhile
\State \Return $H_{1:T}(z^\star)$ \Comment{The policy completes the selected prefix to termination}
\end{algorithmic}
\end{algorithm}

\subsection{CER for Reinforcement Learning}
\label{sec:supervised-rubrics}
\label{sec:rl-method}
Unlike TTS, coding tasks in an RL environment usually provide verifiable rewards. Reusing experience-guided adaptive rubrics would fail to exploit this supervision; moreover, in preliminary experiments we find that adaptive rubrics generated by a relatively weak rubric model provide only weak supervision and cannot match training with test-verifier rewards. We therefore use verifiable rewards to construct task-specific rubrics offline, while retaining verifier-free rubric rewards during online training. This experiment should be viewed as an investigation of whether rubric rewards can be used for long-horizon coding agents when verifiable rewards exist, although the current setup is primarily methodological. With a stronger rubric model, such as a verifier trained specifically for prefix evaluation, the same TTS framework could be applied for RL training.

We first collect historical rollout groups as $\mathcal G_{x,t}=\{(x,h_t^i,s_i)\}_{i=1}^{m}$, including the task $x$, trajectory continuations $h_t^i$ cutoff at a fixed turn budget $t$, and test-verifier rewards $s_i$. A strong teacher model synthesizes task-specific rubrics specifying their applicability, observable evidence, score anchors, and initial rubric weights by comparing different continuations against their verifiable rewards. Candidate rubrics are evaluated by scoring all continuations to obtain rubric rewards $\hat{s}_i=s_x(h_t^i)$, and a generated rubric $j$ is retained only if $\operatorname{Acc}_{\mathrm{pair}}(\hat{\mathbf s}^{(j)},\mathbf s)>\operatorname{Acc}_{\mathrm{pair}}(\hat{\mathbf s}^{(j-1)},\mathbf s)$. We further go through two refinement steps to sharpen ambiguous evidence boundaries and score anchors, where we append the previous rubric's scoring errors to the generation context.

To make rubric rewards reliable for training, we introduce two practical designs. First, we observe that the teacher usually identifies semantically useful criteria, but its initial weights can overemphasize secondary behavior while underweighting the task's central failure mode. We therefore search for task-specific weights $\mathbf w_x^\star$ that maximize pairwise agreement with verifier rewards as detailed in Appendix~\ref{app:weight-optimization}. In our study, simply optimizing the model-assigned weights to $\mathbf w_x^\star$ raises pairwise accuracy from $66.41\%$ to $72.88\%$ on 3,410 held-out pairs. Second, many sampled groups contain no reliable relative learning signal, and the judge noise can be amplified through group normalization. CER first asks the judge for a group-level direct abstention decision $a_{\mathcal G}\in\{0,1\}$, indicating whether the visible prefixes are materially indistinguishable. For groups that do not directly abstain, let $\widetilde r_{ik}\in[0,1]$ be the normalized score of continuation $i$ under rubric $k$. We then computes the within-group score range for each rubric and average:
\begin{equation}
 \Delta_k=\max_i\widetilde r_{ik}-\min_i\widetilde r_{ik},\qquad \Delta_{\mathcal R}=\frac{\sum_{k=1}^{K}w_{x,t}^{k}\Delta_k}{\sum_{k=1}^{K}w_{x,t}^{k}}.
 \label{eq:variance-detector}
\end{equation}
A group is abandoned and resampled if $a_{\mathcal G}=1$ or $\Delta_{\mathcal R}<\theta$. For every retained group, CER samples rollouts only to the fixed turn budget $t$ and uses the centered group advantage $A_i^{\mathrm{CER}}=\hat{s}_i-\frac{1}{m}\sum_{j=1}^{m}\hat{s}_j$, where $\hat{s}_i$ is already normalized to [0, 1]. Therefore, no terminal rollout or test verifier is needed to produce the reward used by the policy update. We optimize these early rollouts using the same token-level binary-TV DPPO objective as TMax \citep{ivison2026tmax}:
\begin{equation}
 \mathcal J_{\mathrm{CER}}(\theta)=\frac{1}{\sum_i |h_t^i|}\sum_{i=1}^{m}\sum_{\ell=1}^{|h_t^i|}m_{i\ell}\,A_i^{\mathrm{CER}}\,\frac{\pi_\theta(h_{t,\ell}^i\mid h_{t,<\ell}^i)}{\pi_{\mathrm{old}}(h_{t,\ell}^i\mid h_{t,<\ell}^i)}
 \label{eq:cer-rl-objective}
\end{equation}
Here $|h_t^i|$ is the number of trainable response tokens, $\pi_{\mathrm{old}}$ is the rollout policy, and $m_{i\ell}$ is the standard DPPO trust-region mask. CER changes only the advantage supplied to this objective.

\section{Experiments}

\begin{table}[t]
\centering
\tabcolsep=0.1cm
\begin{tabular}{lcccc|cccc}
\toprule[1.5pt]
& \multicolumn{4}{c|}{\textsc{Nemotron 3 Ultra}} & \multicolumn{4}{c}{\textsc{Qwen 3.6 27B}} \\
Method & RM@4 & RM@8 & Top-tie & Tok./Ins. & RM@4 & RM@8 & Top-tie & Tok./Ins. \\
 & \% $\uparrow$& \% $\uparrow$ & \% $\downarrow$ & M $\downarrow$& \% $\uparrow$& \% $\uparrow$ & \% $\downarrow$ & M $\downarrow$ \\
\midrule[0.75pt]
SWE-RM &62.2 & 61.6 & 17.8 & 25.67 &66.2 & 67.6 & 18.0 & 5.27 \\
OpenHands Critic@20 &57.6 & 57.8 & 3.2 & \textbf{3.91} &64.4 & 64.2 & 1.6 & \textbf{1.57} \\
OpenHands Critic@40 &60.4 & 58.4 & 2.8 & 5.70 &65.0 & 64.0 & \textbf{0.6} & 2.99 \\
OpenHands Critic@60 &61.8 & 60.4 & 2.4 & 7.93 &65.4 & 63.4 & 1.2 & 4.07 \\
OpenHands Critic@80 &61.4 & 59.6 & \underline{2.0} & 10.25 &66.4 & 64.6 & \underline{1.0} & 4.65 \\
OpenHands Rubrics &59.8 & 58.4 & \textbf{1.4} & 25.57 &63.8 & 62.0 & 1.4 & 5.31 \\
Agentic Rubrics &63.0 & 62.8 & 57.2 & 25.51 &65.6 & 65.6 & 88.8 & 5.36 \\
Adaptive Rubrics &63.0 & 63.4 & 56.8 & 25.95 &69.2 & 69.4 & 31.4 & 5.38 \\
\midrule[0.75pt]
Random &59.0 & 58.6 & - & 3.15 &64.8 & 64.5 & - & 0.64 \\
Best-of-N  & 73.0 & 76.2 & - & 25.19 & 76.4 & 78.6 & - & 5.11 \\
\midrule[0.75pt]
CER@20 &64.4 & 64.8 & 11.8 & \underline{3.96} &67.2 & 68.6 & 3.0 & \underline{1.89} \\
CER@40 &64.4 & \underline{65.2} & 13.1 & 6.28 &69.2 & 70.2 & 8.6 & 3.99 \\
CER@60 &\textbf{66.6} & \textbf{67.6} & 11.2 & 9.14 &\textbf{71.0} & \textbf{71.4} & 13.6 & 5.01 \\
CER@80 &\underline{66.2} & \textbf{67.6} & 12.8 & 11.22 &\textbf{71.0} & \underline{71.0} & 21.1 & 5.34 \\
\bottomrule[1.5pt]
\end{tabular}
\caption{Test-time scaling results on SWE-bench Verified. Top-tie is the fraction of tasks having trajectories receive the equal highest score. Token usage includes online rollout and judging costs. The best results are in \textbf{bold}, and the second-best results are \underline{underlined}, excluding random selection and best-of-N. With only 15.3\% and 74.2\% of the required tokens on Nemotron 3 Ultra and Qwen 3.6 27B, CER matches the best execution-free baseline performance. With no more computation, CER improves upon the best baseline by 4.2 and 2.0 pp on the two models, respectively.}
\label{tab:tts-main}
\end{table}

\begin{table}[t]
\centering
\begin{tabular}{lcccc}
\toprule[1.5pt]
Method & Fold 0 & Fold 1 & Fold 2 & Mean \\
\midrule[0.75pt]
Qwen 3.5 9B & $44.6\,\pm\,1.2$ & $43.1\,\pm\,1.7$ & $45.6\,\pm\,2.7$ & $44.5\,\pm\,0.8$ \\
TMax & $49.7\,\pm\,1.0$ & $49.1\,\pm\,2.7$ & $50.3\,\pm\,2.0$ & $49.7\,\pm\,1.1$ \\
TMax-40 & $49.6\,\pm\,1.6$ & $49.1\,\pm\,2.3$ & $50.7\,\pm\,1.9$ & $49.8\,\pm\,0.8$ \\
CER & $\mathbf{52.7\,\pm\,2.1}$ & $\mathbf{49.8\,\pm\,2.1}$ & $\mathbf{52.4\,\pm\,1.0}$ & $\mathbf{51.6\,\pm\,0.9}$ \\
\bottomrule[1.5pt]
\end{tabular}
\caption{Mean test resolved rate (\%) $\pm$ 95\% CI on the test set, computed from the variance of eight evaluation replicates (one rollout per test instance each); CIs reflect evaluation sampling only, not training variance. CER exceeds full-rollout TMax by 1.9 pp.}
\label{tab:rl-test}
\vspace{-10pt}
\end{table}

\subsection{Models and Dataset} We evaluate on all 500 SWE-bench Verified \citep{jimenez2024swe} instances. For TTS experiments, the experience retriever uses a leave-one-out bank: experience derived from that instance is excluded. For RL, we construct three folds, each partitioning the benchmark into 250 training, 50 validation, and 200 test instances. Training and evaluation splits across three folds are separated at the repository level to the greatest extent apart from one high-volume repository. We use the standard mini-SWE-agent \citep{yang2024sweagent} scaffold for agent--environment interaction and Slime \citep{slime_github} for RL training. For TTS experiments, we evaluate \textsc{Nemotron 3 Ultra} \citep{blakeman2026nemotron} and \textsc{Qwen 3.6 27B} \citep{qwen2026qwen36} with self-generated rubrics. Experiences are summarized by \textsc{GPT-5.6 Sol} \citep{openai2026gpt56}. For RL experiments, we train \textsc{Qwen 3.5 9B} \citep{qwen2026qwen35}; rubrics are generated offline by \textsc{GPT-5.6 Sol}.

\subsection{Baselines}

\paragraph{TTS.} We compare with five baselines, including learned reward models and rubric-based LLM verifiers designed specifically for coding agents.
\textbf{SWE-RM} is an execution-free terminal reward model trained with additional SWE data and predicts whether a completed software-agent trajectory will resolve the task \citep{shum2026swe}.
\textbf{OpenHands Critic} is a critic model trained on additional SWE data with temporal-difference targets propagated from terminal verifier rewards \citep{openhands2025critic}. The critic can predict state values and thus be used for terminal-reward estimation.
\textbf{OpenHands Rubrics} is a fixed behavioral rubric set collected from real-world human--agent interaction trajectories. We use these rubrics for LLM judging without any learned critic or reward model \citep{wang2026rubric}.
\textbf{Agentic Rubrics} generates task-adaptive rubrics. It asks an expert model to inspect the issue and repository, construct a task-specific rubric list, and use it to evaluate completed candidate patches \citep{raghavendra2026agentic}.
\textbf{Adaptive Rubrics} is a strengthened baseline that we construct. It generates task-adaptive rubrics from the issue, repository, summarized trajectory group, and submitted patches. This allows the rubric model to observe the error and success patterns of different rollouts rather than infer them without trajectory evidence, producing more discriminative rubrics. We do not enable the experience bank for this baseline.

\paragraph{RL.} We use exactly the same training configuration as TMax and change only the policy-update horizon and reward. We compare three training setups. \textbf{TMax} trains on full rollouts and final verifier rewards. \textbf{TMax-40} completes each rollout to termination to obtain its final verifier reward but applies policy optimization only to the first 40 steps. \textbf{CER} stops each rollout at 40 steps and uses rubric scores judged only from these prefixes.

\subsection{Evaluation} 

CER is an early reward for unfinished prefixes, and its prefix-level accuracy cannot be measured reliably because a prefix's value is latent and a single completed suffix provides only one Bernoulli sample. We therefore evaluate CER through its downstream TTS task. For TTS, we report RM@$k$, which is the average resolved rate of the rollout with the highest reward among $k$ samples. The selected rollout is $\hat i_x=\arg\max_{1\leq i\leq k}s_x(h_t^i)$. We then compute $\operatorname{RM@}k=\frac{1}{N}\sum_x y(\tau_T^{\hat i_x})$, where $N$ is the number of evaluation tasks. For RL, we report the average resolved rate across $k$ sampled rollouts on the test split, $\operatorname{Resolved}=\frac{1}{Nk}\sum_x\sum_{i=1}^{k}y(\tau_{T,x}^{i})$, using the best performed checkpoint on the validation set. We report 95\% confidence intervals over $k$ samples; these intervals quantify evaluation sampling uncertainty, rather than variation across independent RL training runs.

For TTS, token usage includes all input and output tokens consumed by policy and judge rollouts. For RL, we count all model input and output tokens consumed by non-stale training samples, including filtered groups due to no reward variance. The costs of TMax and TMax-40 include every policy token required to obtain the terminal reward. CER's cost includes policy tokens through turn 40 and all rubric-judge prompt and completion tokens.

\section{Results}

\begin{figure}[t]
\includegraphics[width=0.95\columnwidth]{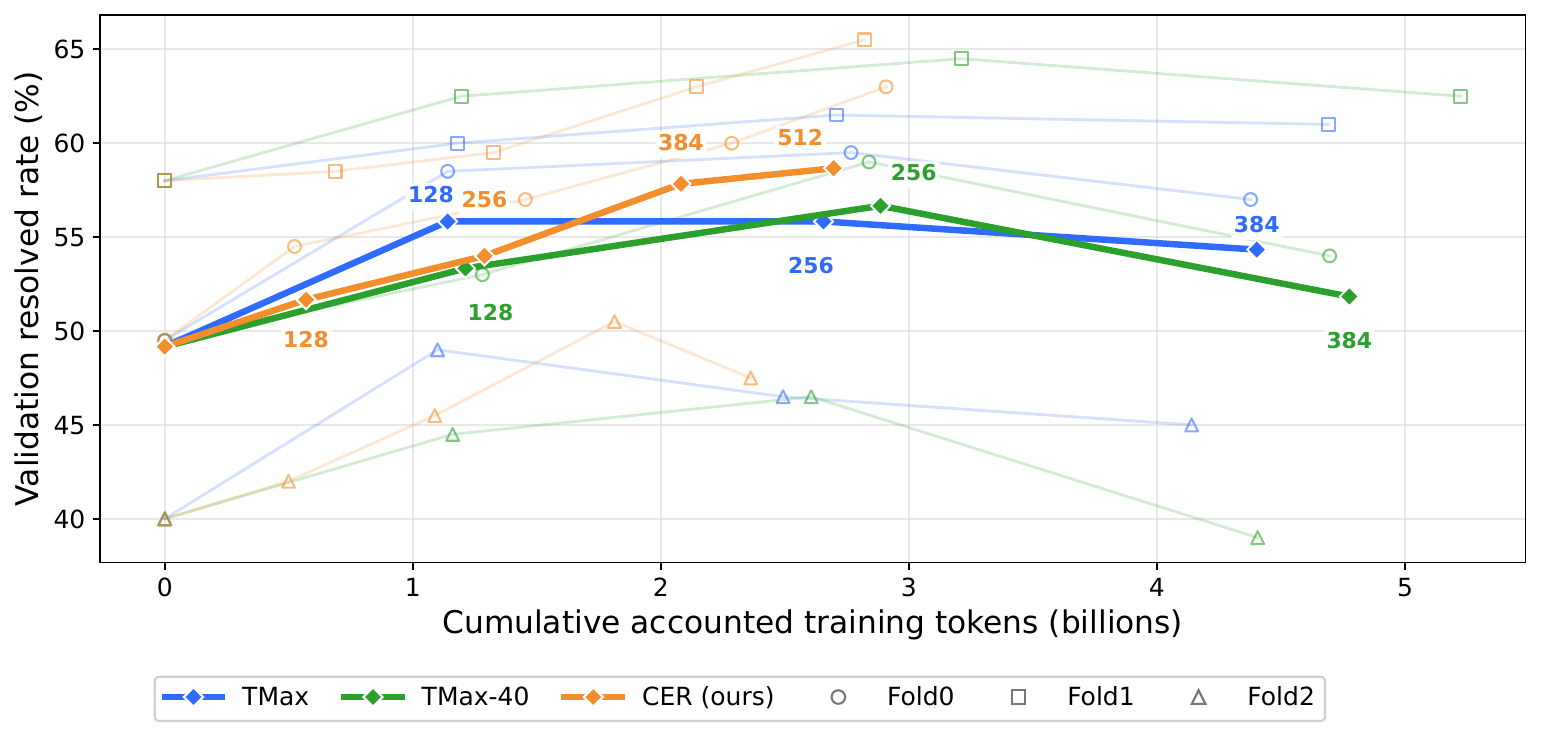}
\caption{Mean resolved rate versus cumulative accounted training tokens across three folds on SWE-bench Verified. Light curves show individual folds, bold curves show their arithmetic means, and numbers indicate the actual trained groups. CER shows a clear trend toward more stable training, lower token usage, and better final performance.}
\label{fig:rl-token-efficiency}
\end{figure}

\subsection{Test-Time Scaling}

Table~\ref{tab:tts-main} reports the TTS results on SWE-bench Verified with various step budgets. CER improves the quality--compute frontier on both models. On Nemotron 3 Ultra, CER with a 20-step budget already exceeds the best complete-trajectory baseline by 1.4 pp on RM@8 while using 84.7\% fewer tokens, and CER@60 reaches the best RM@8 of 67.6\% with 64.8\% fewer tokens. On Qwen 3.6 27B, CER with a 40-step budget improves over the best full-trajectory baseline by 0.8 pp on RM@8 with 25.8\% fewer tokens, while CER@60 reaches the best RM@8 of 71.4\%. Measured against the attainable range between Random selection and the Oracle,  Adaptive Rubrics, our strengthened best baseline, closes 27\% and 35\% of the gap on Nemotron 3 Ultra and Qwen 3.6 27B, whereas CER closes 51\% and 49\%. The advantage persists on the strictly unfinished subsets (Appendix Table~\ref{tab:tts-unfinished-rm}). Across both policies, increasing the prefix budget generally improves CER because longer prefixes expose more evidence about whether the current approach is viable. Qualitative inspection shows that agents' self-correction is usually local rather than hypothesis-revising: they may recover from an unproductive inspection by opening more relevant files, yet rarely abandon an initially incorrect hypothesis about the root cause. This makes early prediction feasible.

We also examine why the baselines have undesirable performance. Generative verifiers that can reason over evidence with chain-of-thought have been shown to outperform numerical reward models in previous work \citep{zhang2025generative}. Generalization is another limitation for trained reward models: SWE-RM is primarily trained on Qwen 3 Coder on-policy data; its larger deficit to the best CER result on Nemotron 3 Ultra than on Qwen 3.6 27B (6.0 versus 3.8 pp) is therefore consistent with policy and scaffold shift. Rubric-based methods are more generalizable, but their criteria can remain too coarse to capture behavioral differences within candidate groups. We quantify this using the top-tie rate, the fraction of groups in which multiple trajectories receive the maximum score. Agentic Rubrics ties on 57.2\% of Nemotron groups and 88.8\% of Qwen groups, whereas CER@20 ties on only 11.8\% and 3.0\%, respectively. This gap indicates that prior rubric construction often remains too coarse to distinguish trajectories, while CER's group-conditioned rubric refinement resolves substantially more of the behavioral differences.

\subsection{Reinforcement Learning}

\begin{wraptable}{r}{0.50\columnwidth}
\centering
\vspace{-15pt}
\scriptsize
\resizebox{\linewidth}{!}{\begin{tabular}{lrrrr}
\toprule[1.5pt]
CER & @20 & @40 & @60 & @80 \\
\midrule[0.75pt]
Nemotron & 63\% & 68\% & 68\% & 69\% \\
$-$ w/ Qwen exp. & 64\% & 69\% & 70\% & 70\% \\
\midrule[0.75pt]
Qwen & 73\% & 75\% & 76\% & 75\% \\
$-$ w/ Nemotron exp. & 67\% & 68\% & 69\% & 70\% \\
\bottomrule[1.5pt]
\end{tabular}}
\caption{Cross-model experience transfer on the same 100-instance subset. Entries are resolved RM@8 (\%) at each budget $k$. Experience from the stronger model improves the weaker model's TTS performance, whereas the reverse transfer is detrimental.}
\label{tab:tts-transfer}
\vspace{-10pt}
\end{wraptable}

Figure~\ref{fig:rl-token-efficiency} reports validation resolved rate against cumulative training-token cost across three folds, and Table~\ref{tab:rl-test} reports test resolved rate for the checkpoint selected on each validation fold. At the matched 384-step checkpoint, CER reaches 57.8\% validation resolved rate, 3.5 pp above TMax while using 52.7\% fewer online policy-and-judge tokens during RL. Extending CER to 512 steps raises validation performance to 58.7\% while still requiring equivalent token budget with TMax's 256 steps. On the test sets, CER exceeds full-rollout TMax by 1.9 pp. The online token saving comes from both shorter trajectories and denser supervision. Through the matched 384-step checkpoint, accepted TMax has average 60.2 assistant turns per trajectory, whereas CER averages 37.3. TMax filters or resamples 34.3\% groups because terminal rewards have zero variance; CER abstains only 9.9\%. Thus CER both reduces the cost per trajectory and extracts useful comparisons from more sampled groups, even after accounting for judge tokens.

Prefix-only optimization is effective because policy uncertainty is concentrated in the early steps of the trajectory. Mean token entropy is 0.399 nats over turns 1--40 and 0.312 nats over turns 41--80, as shown in Appendix Figure~\ref{fig:rl-rollout-entropy}. Later actions are more constrained by the accumulated trajectory state, so the prefix contains a larger share of the policy's uncertainty. This observation is consistent with PivotRL, which restricts updates to high-impact intermediate turns while reducing rollout computation \citep{yi2026pivotrl}. Although both TMax-40 and CER use prefix-only optimization, terminal verifiable rewards prove less effective. The terminal verifier is more likely to assign identical rewards to every trajectory, so TMax-40 cannot learn from zero-variance groups, whereas distinguishable behavioral differences remain visible in the prefixes even when the final outcomes are identical. Moreover, rubrics provide more interpretable rewards grounded only in prefix behaviors, while terminal rewards may mix information about future actions or unobserved terminal outcomes. These prefix-unobservable signals act as reward noise and can impede effective learning.

Our TMax run is a controlled baseline rather than a reproduction of the original final model: TMax trains on 14,600 synthetic tasks, whereas each of our SWE-bench folds trains on only 250 tasks. In this data-limited setting, TMax reaches its best mean validation rate by 128-256 steps and quickly saturates, so its best-validation checkpoint already reflects its peak performance. CER provides a more data-efficient use of this limited or saturated task pool with denser reward signals.

\section{Analysis}

\subsection{Experience Bank Transferability Across Models}
\label{sec:transfer}

Table~\ref{tab:tts-transfer} shows the TTS performance of Nemotron 3 Ultra and Qwen 3.6 27B on a sampled 100-instance SWE-bench Verified subset when exchanging their experience banks. To separate experience-content effectiveness from retrieval accuracy, we replace the source-model experience bank's retrieval keywords with the target model's keywords wherever source tasks align. Qwen-derived experience improves Nemotron at every round by 1--2 pp, while Nemotron-derived experience reduces Qwen by 5--7 pp. This demonstrates that transferability is directional: experience learned from a stronger model can benefit a weaker model's TTS more than its own experience bank, whereas experience from a weaker model harms a stronger model by a substantially larger margin, showing that cross-model transfer is asymmetric rather than proportional.

\subsection{Rubric Judge Agreement with Verifiable Reward}
\label{sec:judge-analysis}

Appendix Figure~\ref{fig:rl-pairwise-accuracy} tracks pairwise agreement between rubric rewards and verifier rewards throughout the CER runs reported in Figure~\ref{fig:rl-token-efficiency}. Across 21,938 such pairs, overall pairwise agreement is 60.9\%, and the average ranges from 53.9\% to 70.1\% over training. This moderate agreement shows that CER does not merely reproduce the terminal verifier: it induces a related but distinct ordering over prefix behaviors. The disagreement may arise from rubric-judge errors, meaningful intermediate behaviors that are not reflected by terminal tests, or verifier outcomes determined by unseen suffixes. This online agreement is not directly comparable to the offline held-out accuracy in Section~\ref{sec:rl-method}, which is measured on the historical rollouts used to construct the rubrics; online rollouts additionally drift from that distribution as the policy is updated.

\subsection{Qualitative and Quantitative Analysis of Rubrics}
\label{sec:qualitative}

A quantitative analysis of 1,891 generated rubrics across 500 tasks focuses primarily on three stages: 22.5\% evaluate exploration and diagnosis, 48.2\% evaluate implementation and editing, 29.0\% evaluate validation, and only 0.3\% concern other nonfunctional risks. Qualitatively, different tasks receive different stage emphases. In 89.2\% of tasks, one stage receives more than half of the total weight, and the median dominant-stage share is 74.2\%. Even the same rubric category is instantiated differently across tasks. For Django-14017, efficient exploration means using each contradictory Q-expression probe to eliminate the mechanism it disproves and narrow the hypothesis; for scikit-learn-14053, it means extracting a concrete source constraint from blocked execution instead of continuing environment setup or cycling through incompatible index changes. These task-specific criteria direct the judge toward small but diagnostic evidence that a generic rubric could overlook. Qualitative results demonstrate that CER generates highly adaptive rubrics. Appendix Tables~\ref{tab:rubric-taxonomy-exploration}--\ref{tab:rubric-taxonomy-validation} provide the primary clusters and examples.

\section{Conclusion}
We study whether long-horizon software-agent trajectories can be evaluated before they terminate and whether such early rewards are useful beyond prediction itself. CER constructs task-conditioned rubrics from retrieved historical experience and sampled rollouts within a group; these rubrics can then be used for downstream test-time search and terminal-free group-relative RL. Across the Nemotron 3 Ultra and Qwen 3.6 27B TTS experiments, CER improves RM@8 by 4.2 and 2.0 pp over the best baselines and saves 84.7\% and 25.8\% of tokens, respectively, at matching performance. In the three-fold Qwen 3.5 9B RL experiment, CER exceeds full-rollout TMax by a small margin (51.6\% versus 49.7\%) while using 52.7\% fewer online policy-and-judge tokens during RL. This work revisits whether rubric-based LLM evaluators remain valuable in a verifiable environment and demonstrates their effectiveness for early terminal-reward prediction. As test-based final verifiable rewards become increasingly insufficient and sparse for long-horizon agents, we hope the proposed rubric design and analysis can inspire future work on behavior-focused data selection, efficient and reflective test-time scaling, and RL training with rubric rewards for coding agents.

\subsection*{AI use statement}
We used generative AI tools for tasks requiring disclosure: generating experimental trajectories, experiences, rubrics, and related synthetic artifacts;  formulating mathematical claims and assisting with proofs; implementing and debugging methods; processing data; supporting qualitative analysis; and interpreting experimental results. We did not use generative AI for developing and refining the conceptual framework, hypotheses, methodology, and experimental design. We additionally used generative AI for recommended-disclosure tasks, including suggesting experimental parameters, creating and editing figures, searching and identifying relevant work, drafting and editing paper text, improving readability and presentation. We manually verified cited sources, tested AI-assisted code, audited reported results against saved results, and reviewed all AI-assisted text and analyses. The language models studied as policies, rubric generators, judges, and skill summarizers are part of the research method and are documented separately in the paper. The authors take responsibility for the final content, claims, code, and artifacts.

\subsection*{Ethics statement}
This work uses public software-engineering benchmarks derived from open-source repositories and does not involve human subjects or the collection of private user data. Agent execution and test evaluation are isolated in sandboxed environments. Automated coding and evaluation systems may nevertheless produce incorrect or insecure patches, amplify biases in benchmark and task selection, or be misapplied without human oversight; rubric judgments are also imperfect and should not replace security review or maintainer judgment. We therefore recommend human review before deployment.

\subsection*{Reproducibility statement}
The main text specifies CER's formal objective, algorithms, evaluation protocols, data splits, models, and principal training settings. The appendix provides complete proofs, robust rubric-weight optimization, repository-level split statistics, full RL hyperparameters, retrieval and score-aggregation details, prompts, and training dynamics. 
Codes for reproducing our experiments are available at this link: \href{https://github.com/yaojh18/RLER}{https://github.com/yaojh18/RLER}.

\newpage

\bibliography{iclr2027_conference}
\bibliographystyle{iclr2027_conference}

\newpage
\appendix

\begin{wrapfigure}{r}{0.50\columnwidth}
\centering
\vspace{-10pt}
\includegraphics[width=\linewidth]{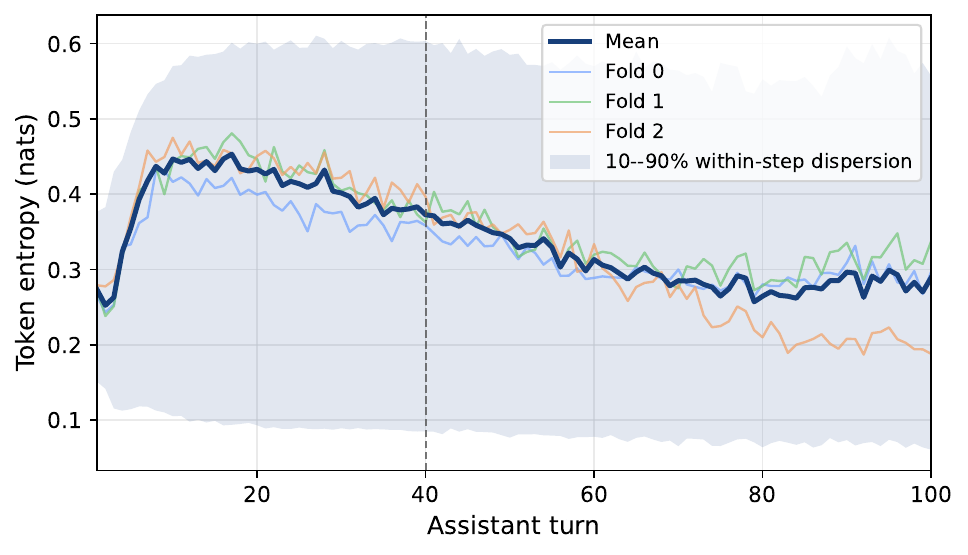}
\caption{Token entropy across the first 100 turns of TMax training rollouts. Light lines show fold means and the dark line shows the pooled mean; shading denotes the 10th--90th percentile across trajectory-level turn means within each turn. The dashed line marks CER's 40-turn cutoff.}
\label{fig:rl-rollout-entropy}
\vspace{-10pt}
\end{wrapfigure}

\section{Limitations}
CER does not eliminate the cost of constructing its experience and supervised rubrics: the historical trajectories must first be collected and terminally evaluated. This is an up-front investment rather than a per-query requirement because the resulting bank is reused across subsequent tasks and stages. Its benefit therefore grows with reuse, and our generalization studies demonstrate that experience summarized from stronger models can be used to guide weaker models.

CER also introduces a synchronous judging stage into TTS. Let $k$ be the number of rollouts, $N$ the step limit, $R$ the number of search rounds, and $j$ the latency of one judge call. A conventional generate-then-rerank pipeline has critical-path cost $O(kN+j)$ under sequential execution, whereas CER has $O(kN+Rj)$ because generation pauses for rubric-guided decisions. In our setting, $j$ is much smaller than a long agent rollout, but these pauses remain a systems bottleneck. With parallel rollout generation and pipelining between ready branches, both pipelines can approach an $O(N)$ rollout-dominated critical path; realizing this benefit consistently requires better asynchronous scheduling infrastructure.

\subsection{Offline Rubric \& Experience Bank Construction Cost}
Let $k$ be the number of historical rollouts, $m$ the number of rubric-generation and refinement passes, and $C_{\mathrm{roll}}$ and $C_{\mathrm{judge}}$ the costs of one rollout and one rubric/experience generation-plus-judging pass, respectively. Ignoring CPU-only weight optimization, CER's one-time construction cost is
\begin{equation}
 C_{\mathrm{CER}}=kC_{\mathrm{roll}}+mkC_{\mathrm{judge}}.
 \label{eq:cer-offline-cost}
\end{equation}
A reward model trained for $e$ epochs, with per-rollout training cost $C_{\mathrm{train}}$, has the corresponding cost
\begin{equation}
 C_{\mathrm{RM}}=kC_{\mathrm{roll}}+ekC_{\mathrm{train}}.
 \label{eq:rm-offline-cost}
\end{equation}
When $m=e$ and $C_{\mathrm{judge}}\leq C_{\mathrm{train}}$, CER's construction cost is no greater than training a reward model under this accounting. We use $m=3$ passes in our experiments, which is usually less than the typical training epoch required for SFT. The inequality depends on the teacher model's inference cost and the reward model's training cost; for the same model, judging is inference-only and is therefore ordinarily cheaper than training on the same rollout.

\section{Related Work (Cont')}

\paragraph{Experience reuse and memory.} Memory-augmented agents reuse prior interactions as demonstrations, reflections, or summarized skills. SkillRL, for example, distills trajectories into a hierarchical skill bank and retrieves skills or structured playbooks of strategies from previous execution feedback to guide the acting policy \citep{xia2026skillrl, zhang2026agentic}. These methods use accumulated experience to improve solution generation. ARBOR instead maintains a reusable rubric buffer summarized from contrasting trajectories and adds the resulting process scores to outcome rewards for multi-hop QA tasks \citep{liu2026arbor}; its short-horizon trajectories are scored after completion, so predicting the terminal reward of an unfinished prefix does not arise. We likewise use experience for evaluation, but for long-horizon coding agents before the rollout ends: retrieved experience defines which judging criteria worked well for historical tasks, helping rubric models generate the most discriminative rubrics for the current task.

\section{Theoretical Proofs}
\label{app:advantage-bound}

\subsection{Offset Invariance and Order Invariance for TTS}

Let $s_i=s_x(h_t^{(i)})$ and $v_i=V_{x,\pi}(h_t^{(i)})$ for a non-degenerate group. If $s_i=v_i+b_H$ for a group-specific offset $b_H$, mean centering removes the offset exactly:
\begin{equation}
 s_i-\bar s=v_i-\bar v.
 \label{eq:offset-invariance}
\end{equation}
TTS requires only the induced order. Let $\operatorname{Top}_k(r)$ denote the indices of the $k$ largest entries of a score vector $r$. Under Eq.~\ref{eq:ranking}, whenever the $k$-th and $(k+1)$-th ideal values are distinct,
\begin{equation}
 \operatorname{Top}_k(s)=\operatorname{Top}_k(v).
 \label{eq:topk-invariance}
\end{equation}
Suppose instead that some $i\in\operatorname{Top}_k(v)$ were replaced by $j\notin\operatorname{Top}_k(v)$ under $s$. The strict boundary gives $v_i>v_j$, whereas the replacement requires $s_j>s_i$, contradicting Eq.~\ref{eq:ranking}. If ideal values tie at the boundary, exchanging tied candidates leaves the rank-based objective unchanged.

\subsection{Pairwise Form of a DPPO Update}

Let $r_i$ be any reward assigned to trajectory $i$, let $\bar r=m^{-1}\sum_i r_i$, and define the token-level importance ratio
\begin{equation}
 \rho_{i\ell}(\theta)=\frac{\pi_\theta(h_{t,\ell}^i\mid h_{t,<\ell}^i)}{\pi_{\mathrm{old}}(h_{t,\ell}^i\mid h_{t,<\ell}^i)}.
 \label{eq:dppo-importance-ratio}
\end{equation}
For the DPPO mask $m_{i\ell}^{r}(\theta)$ induced by the sign of $r_i-\bar r$, collect the retained importance-weighted token gradients of trajectory $i$ as
\begin{equation}
 g_i^{r}(\theta)=\sum_{\ell=1}^{|h_t^i|}m_{i\ell}^{r}(\theta)\rho_{i\ell}(\theta)\nabla_\theta\log\pi_\theta(h_{t,\ell}^i\mid h_{t,<\ell}^i).
 \label{eq:dppo-weighted-trajectory-gradient}
\end{equation}
The actual DPPO update induced by $r$, up to the common token normalization $Z=\sum_i|h_t^i|$, is
\begin{equation}
 G_\theta(r)=\frac{1}{Z}\sum_{i=1}^{m}(r_i-\bar r)g_i^{r}(\theta).
 \label{eq:centered-dppo-gradient}
\end{equation}
The same algebraic identity gives the exact pairwise form
\begin{equation}
 G_\theta(r)=\frac{1}{mZ}\sum_{i<j}(r_i-r_j)\bigl(g_i^{r}(\theta)-g_j^{r}(\theta)\bigr).
 \label{eq:pairwise-dppo-gradient}
\end{equation}

\begin{wrapfigure}{r}{0.50\columnwidth}
\centering
\vspace{-20pt}
\includegraphics[width=\linewidth]{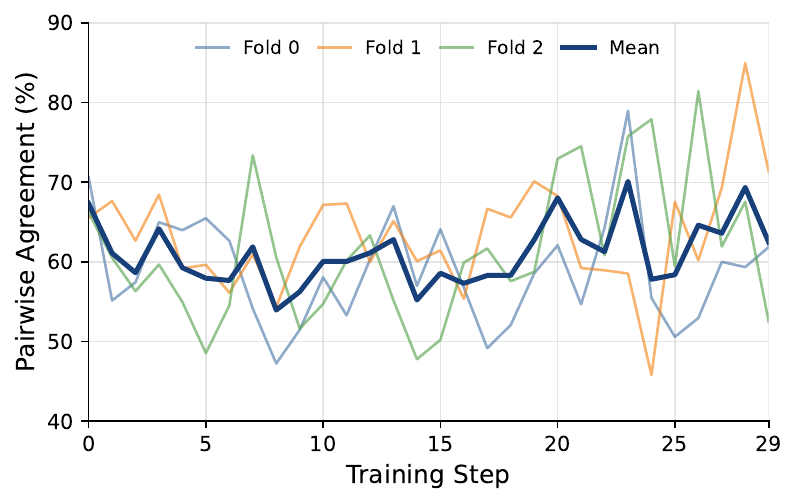}
\caption{Rubric--verifier pairwise agreement throughout the CER runs used in Figure~\ref{fig:rl-token-efficiency}. Light lines show individual folds and the dark line is the pair-count-weighted mean. Only pairs with different terminal rewards are included; rubric-score ties count as incorrect.}
\label{fig:rl-pairwise-accuracy}
\vspace{-30pt}
\end{wrapfigure}

Equation~\ref{eq:pairwise-dppo-gradient} shows that reward-gap distortion affects a DPPO update by reweighting pairwise gradient differences. Pairwise order consistency guarantees that, for each pair, the reward-gap coefficient preserves its sign; therefore, when the corresponding gradient difference is fixed, reward-gap distortion can only rescale rather than reverse that pairwise contribution. It does not, however, control cancellations among contributions from different pairs. We next state an additional sufficient batch-level stability condition.

\subsection{A Batch-Stability Requirement for Rubric Rewards}

Pairwise order consistency is sufficient for rank-based TTS, but it cannot by itself guarantee that rubric rewards reproduce the numerical gaps used by DPPO. We therefore state an additional sufficient property for an ideal RL rubric. For $B$ independently sampled groups in an optimizer batch, let $G_{\theta,b}(s)$, $G_{\theta,b}(v)$ denote the actual DPPO update from the $b$-th group under rubric score $s$ and verifiable reward $r$, as defined in Eq.~\ref{eq:centered-dppo-gradient}. We require the rubric scores to be unbiased across groups:
\begin{equation}
 \mathbb E\bigl[G_{\theta,b}(s)-G_{\theta,b}(v)\bigr]=0.
 \label{eq:batch-stable-unbiasedness}
\end{equation}
Assume rubric-induced reward distortion is bounded by $\epsilon$:
\begin{equation}
 \mathbb E\bigl[\lVert G_{\theta,b}(s)-G_{\theta,b}(v)\rVert_2^2\bigr]\leq\epsilon^2\mathbb E\bigl[\lVert G_{\theta,b}(v)\rVert_2^2\bigr].
 \label{eq:bounded-relative-group-distortion}
\end{equation}
For two different independently sampled groups, Eq.~\ref{eq:batch-stable-unbiasedness} gives
\begin{equation}
 \mathbb E\!\left[\left\langle G_{\theta,b}(s)-G_{\theta,b}(v),G_{\theta,c}(s)-G_{\theta,c}(v)\right\rangle\right]=0,\qquad b\neq c.
 \label{eq:cross-group-error}
\end{equation}
Expanding the squared batch-average error and applying Eq.~\ref{eq:cross-group-error} yields
\begin{equation}
 \mathbb E\!\left[\left\lVert\frac{1}{B}\sum_{b=1}^{B}\bigl(G_{\theta,b}(s)-G_{\theta,b}(v)\bigr)\right\rVert_2^2\right]=\frac{1}{B^2}\sum_{b=1}^{B}\mathbb E\bigl[\lVert G_{\theta,b}(s)-G_{\theta,b}(v)\rVert_2^2\bigr].
 \label{eq:batch-update-mse-expansion}
\end{equation}
Applying Eq.~\ref{eq:bounded-relative-group-distortion} gives
\begin{equation}
 \mathbb E\!\left[\left\lVert\frac{1}{B}\sum_{b=1}^{B}\bigl(G_{\theta,b}(s)-G_{\theta,b}(v)\bigr)\right\rVert_2^2\right]\leq\frac{\epsilon^2}{B}\left(\frac{1}{B}\sum_{b=1}^{B}\mathbb E\bigl[\lVert G_{\theta,b}(v)\rVert_2^2\bigr]\right).
 \label{eq:batch-update-mse}
\end{equation}
Provided the normalizing scale below is nonzero, Markov's inequality gives, for any $\delta\in(0,1)$,
\begin{equation}
 \Pr\!\left(\frac{\left\lVert B^{-1}\sum_{b=1}^{B}\bigl(G_{\theta,b}(s)-G_{\theta,b}(v)\bigr)\right\rVert_2}{\sqrt{B^{-1}\sum_{b=1}^{B}\mathbb E\bigl[\lVert G_{\theta,b}(v)\rVert_2^2\bigr]}}\leq\frac{\epsilon}{\sqrt{B\delta}}\right)\geq1-\delta.
 \label{eq:batch-update-concentration}
\end{equation}
In plain words, $\epsilon$ measures how much score distortion the rubric introduces on one group relative to a verifiable reward. Equation~\ref{eq:batch-update-concentration} states that averaging $B$ groups reduces the non-systematic part of this distortion by a factor of $\sqrt{B}$. Individual rubric errors therefore cannot produce an equally large optimizer error, provided that they do not repeatedly push the policy in the same wrong direction. Pairwise accuracy alone does not guarantee this property: an ideal RL rubric must also avoid systematic directional bias across groups, because such bias will not disappear as the batch grows.

\section{Robust Rubric Weight Optimization}
\label{app:weight-optimization}

Rubric generation determines the task-specific criteria and their scoring anchors, but the teacher-assigned weights are not fitted to rollout outcomes and can overemphasize secondary behaviors. For each task $x$, weight optimization takes the accepted rubric set $\mathcal R_{x,t}$, the collected historical groups $\mathcal G_{x,t}=\{(x,h_t^i,s_i)\}_{i=1}^{m}$, the criterion-level judge scores $r_{x,t}^{k}(h_t^i)$, and the teacher weights as input. The rubric text, anchors, polarity, and scoring rule remain fixed throughout optimization.

To make optimized weights robust to judging noise, we deliberately add perturbations $c$ to the rubric scores. Let $r_{x,t}^{k,c}(h_t^i) = r_{x,t}^{k}(h_t^i) + c$ and $\hat s_i^{c}(\mathbf w_x)=\frac{\sum_{k=1}^{K}w_x^k r_{x,t}^{k,c}(h_t^i)}{\sum_{k=1}^{K}w_x^k}$ be the aggregate rubric rewards produced by weights $\mathbf w$ under perturbation $c$. We optimize the average pairwise agreement across these perturbations:
\begin{equation}
 \bar A(\mathbf w)=\frac{1}{C}\sum_{c=1}^{C}\operatorname{Acc}_{\mathrm{pair}}\!\left(\hat{\mathbf s}^{c}(\mathbf w),\mathbf s\right).
 \label{eq:robust-pairwise-objective}
\end{equation}
The weight-optimization problem is therefore
\begin{equation}
 \mathbf w_x^\star=\arg\max_{\mathbf w}\bar A(\mathbf w).
 \label{eq:robust-weight-objective}
\end{equation}
Let $\Omega_w$ denote the allowed values for one rubric weight, $B$ the maximum number of nonzero weights, and $\Omega_{\mathrm{swap}}(w_i)$ the allowed weights for a newly activated rubric when rubric $i$ is removed. Let $\{\mathbf w^{(n)}\}_{n=1}^{N}$ be the initial weight vectors and $\mathbf e_i$ the $i$-th coordinate vector. Algorithm~\ref{alg:robust-weight-optimization} uses only the robust objective $\bar A$; perturbation construction is independent of the search.

We use multiple initial vectors to reduce sensitivity to local optima; examples include retaining the $B$ rubrics with the largest teacher-assigned weights and setting all remaining weights to zero, single-rubric vectors, equal-weight combinations, and random vectors, from which the $N$ highest-$\bar A$ candidates are retained. In our experiments, $\Omega_w=\{0,0.5,0.6,\ldots,5.0\}$, $B=6$, $N=12$, $N_1=5$, and $N_2=2$. For swaps, $\Omega_{\mathrm{swap}}(w_i)=\{0.5,1.0,2.0,\ldots,5.0,w_i\}$. We use $C=200$ sets of perturbed scores: each original five-level score is decreased by one with $10\%$ probability, unchanged with $80\%$ probability, and increased by one with $10\%$ probability, with clipping to $[1,5]$.

In our experiments, simply optimizing the model-assigned weights to $\mathbf w_x^\star$ on 13,639 trained pairs raises pairwise accuracy from $66.41\%$ to $72.88\%$ on 3,410 held-out pairs with different terminal rewards. The split is at the pair level: the historical trajectories are shared, and only their pairs are partitioned into training and held-out sets.

\begin{algorithm}[t]
\caption{Rubric Weight Optimization}
\label{alg:robust-weight-optimization}
\begin{algorithmic}[1]
\Require Robust objective $\bar A$, initial weights $\{\mathbf w^{(n)}\}_{n=1}^{N}$, available weight sets $\Omega_w$ and $\Omega_{\mathrm{swap}}$, iteration limits $N_1$ and $N_2$
\For{$n=1,\ldots,N$}
\State $\mathbf w\gets\mathbf w^{(n)}$
\For{$q=1,\ldots,N_1$}
\For{$k=1,\ldots,K$} \Comment{$K$ is the number of rubric criteria}
\State $w_k\gets\arg\max_{a\in\Omega_w}\bar A(w_1,\ldots,w_{k-1},a,w_{k+1},\ldots,w_K)$ 
\EndFor
\If{no weight changes}
\State \textbf{break}
\EndIf
\EndFor
\For{$q=1,\ldots,N_2$}
\State $(i^\star,j^\star,a^\star)\gets\arg\max_{i:w_i>0,\,j:w_j=0,\,a\in\Omega_{\mathrm{swap}}(w_i)}\bar A(\mathbf w-w_i\mathbf e_i+a\mathbf e_j)$
\If{$\bar A(\mathbf w-w_{i^\star}\mathbf e_{i^\star}+a^\star\mathbf e_{j^\star})\leq\bar A(\mathbf w)$}
\State \textbf{break}
\EndIf
\State $\mathbf w\gets\mathbf w-w_{i^\star}\mathbf e_{i^\star}+a^\star\mathbf e_{j^\star}$
\EndFor
\State $\mathbf w^{(n)}\gets\mathbf w$
\EndFor
\State \Return $\mathbf w_x^\star\gets\arg\max_{\mathbf w\in\{\mathbf w^{(1)},\ldots,\mathbf w^{(N)}\}}\bar A(\mathbf w)$
\end{algorithmic}
\end{algorithm}

\section{Experiments Details}

\subsection{Models, Checkpoints, and Data}
\label{app:models-data}

We employ five LLMs in the experiments, as detailed below:
\begin{itemize}[leftmargin=*]
    \item \textsc{Nemotron 3 Ultra}, through the \href{https://huggingface.co/nvidia/NVIDIA-Nemotron-3-Ultra-550B-A55B-BF16}{\textsc{nvidia/NVIDIA-Nemotron-3-Ultra-550B-A55B-BF16}} checkpoint on Hugging Face \citep{wolf2020transformers}, used for TTS policy rollouts, rubric generation, and judging.
    \item \textsc{Qwen 3.6 27B}, through the \href{https://huggingface.co/Qwen/Qwen3.6-27B}{\textsc{Qwen/Qwen3.6-27B}} checkpoint on Hugging Face, used for TTS policy rollouts, rubric generation, and judging.
    \item \textsc{Qwen 3.5 9B}, through the \href{https://huggingface.co/Qwen/Qwen3.5-9B}{\textsc{Qwen/Qwen3.5-9B}} checkpoint on Hugging Face, used as the RL base model and rollout policy.
    \item \textsc{GPT-5.6 Sol}, through the \textsc{gpt-5.6-sol} checkpoint on OpenAI API, used for TTS experience and RL rubrics generation.
    \item \textsc{GPT-5.6 Luna}, through the \textsc{gpt-5.6-luna} checkpoint on OpenAI API, used for RL rubric judging. \textsc{GPT-5.6 Luna} is a low-cost model whose per-token price is comparable to serving \textsc{Qwen 3.5 9B}.
\end{itemize}

We enable thinking mode for all open-sourced models in TTS and RL. For RL, we separate model-development and test repositories to the greatest extent permitted by the benchmark; the sole exception is \path{django/django}, whose 231 instances are too numerous to place in a single fold. Table~\ref{tab:rl-repository-splits} reports the exact repository distribution; the parenthesized values are instance counts. During training, we sample eight rollouts per task with temperature 1.0 and top-$p$ 0.95, using a context length of 65,536 tokens and a maximum completion length of 20,480 tokens. The rest of the training configuration is exactly the same as TMax. Validation evaluations sample four rollouts per task, whereas test evaluations sample eight, both with temperature 0.7 and a context length of 128,000 tokens. In the results reported in Table \ref{tab:tts-main} and Table \ref{tab:rl-test}, we report the resolved /unresolved rate given by the official verifier. During RL training, we follow the original setup for TMax and use passed test rate as final rewards, which is also continuous and comparable to CER rewards.

The baseline methods are evaluated on exactly the same set of complete trajectory candidates: eight rollouts sampled independently per task. Random reports the expected resolved rate of a uniformly chosen candidate, Best-of-N is the oracle pass@8, and OpenHands Critic@$k$ scores the first $k$ steps of the same shared rollouts and directly selects highest-scored trajectory without any search. Since CER requires beam parent expansion unlike the random sampling of baselines, it is evaluated on candidate prefixes, with each search round extending the best beam parents by 20 agent steps. CER@$k$ denotes a total budget of $k$ agent steps, i.e., $k/20$ search rounds, each of which samples $m=8$ continuations split evenly across the beam parents. At budget $k$, CER reports the continuation selected in round $k/20$ by Algorithm~\ref{alg:cer-tts}; a fully regressed round keeps the previous selection, and the selected prefix is completed by the policy to obtain its terminal reward. The token cost of Openhands Critic and CER comes from the prefix group rollouts, selected prefixed continuation to terminal and judge tokens. For other methods, the token cost includes complete trajectory group rollouts and judge tokens.

\begin{table*}[t]
\centering
\scriptsize
\setlength{\tabcolsep}{4pt}
\renewcommand{\arraystretch}{1.3}
\begin{tabular}{@{}cp{0.3\textwidth}p{0.3\textwidth}p{0.3\textwidth}@{}}
\toprule[1.5pt]
Fold & Train (250) & Validation (50) & Test (200) \\
\midrule[0.75pt]
0 & \path{django/django} (103), \path{matplotlib/matplotlib} (27), \path{psf/requests} (4), \path{pydata/xarray} (16), \path{sphinx-doc/sphinx} (36), \path{sympy/sympy} (64) & \path{django/django} (13), \path{matplotlib/matplotlib} (7), \path{pallets/flask} (1), \path{psf/requests} (4), \path{pydata/xarray} (6), \path{sphinx-doc/sphinx} (8), \path{sympy/sympy} (11) & \path{astropy/astropy} (22), \path{django/django} (115), \path{mwaskom/seaborn} (2), \path{pylint-dev/pylint} (10), \path{pytest-dev/pytest} (19), \path{scikit-learn/scikit-learn} (32) \\
1 & \path{django/django} (102), \path{pydata/xarray} (16), \path{pylint-dev/pylint} (6), \path{scikit-learn/scikit-learn} (25), \path{sphinx-doc/sphinx} (36), \path{sympy/sympy} (65) & \path{django/django} (13), \path{mwaskom/seaborn} (2), \path{pydata/xarray} (6), \path{pylint-dev/pylint} (4), \path{scikit-learn/scikit-learn} (7), \path{sphinx-doc/sphinx} (8), \path{sympy/sympy} (10) & \path{astropy/astropy} (22), \path{django/django} (116), \path{matplotlib/matplotlib} (34), \path{pallets/flask} (1), \path{psf/requests} (8), \path{pytest-dev/pytest} (19) \\
2 & \path{astropy/astropy} (14), \path{django/django} (207), \path{matplotlib/matplotlib} (24), \path{pylint-dev/pylint} (5) & \path{astropy/astropy} (8), \path{django/django} (24), \path{matplotlib/matplotlib} (10), \path{mwaskom/seaborn} (2), \path{pallets/flask} (1), \path{pylint-dev/pylint} (5) & \path{psf/requests} (8), \path{pydata/xarray} (22), \path{pytest-dev/pytest} (19), \path{scikit-learn/scikit-learn} (32), \path{sphinx-doc/sphinx} (44), \path{sympy/sympy} (75) \\
\bottomrule[1.5pt]
\end{tabular}
\caption{Repository and instance distribution of the three best-effort repository-separated RL folds. Fold 2 is fully repository-disjoint; Django is the only repository shared between test and model-development data in Folds 0 and 1.}
\label{tab:rl-repository-splits}
\end{table*}

\subsection{Full RL Training Hyperparameters}

We follow the TMax optimization recipe. Table~\ref{tab:rl-hyperparameters} reports the effective configuration.

\begin{table*}[t]
\centering
\small
\setlength{\tabcolsep}{5pt}
\begin{tabular}{@{}p{0.15\textwidth}p{0.4\textwidth}p{0.4\textwidth}@{}}
\toprule[1.5pt]
Component & Hyperparameter & Value \\
\midrule[0.75pt]
\multirow{7}{*}{Rollout} & Samples per task group & 8 \\
& Samples per global batch & 128 \\
& Sampling temperature / top-$p$ & 1.0 / 0.95 \\
& Context / Completion maximum length & 65,536 / 20,480 tokens\\
& Rollout step limit & 250 (TMax and TMax-40); 40 (CER) \\
& Dynamic sampling & true \\
& Maximum non-variance group retry & 4 \\
\midrule 
\multirow{8}{*}{Optimization} & Objective & DPPO with per-token loss \\
& Group advantage & No standard-deviation normalization \\
& Optimizer & AdamW \\
& Learning-rate schedule & Constant $1\times10^{-6}$; zero warmup \\
& Adam $\beta_1,\beta_2,\epsilon$ & $0.9,0.999,10^{-8}$ \\
& Weight decay & 0 \\
& Gradient clipping & 1.0 \\
& KL coefficients & 0 \\
& Entropy coefficients & 0 \\
& DPPO total-variation threshold & 0.1 \\
\midrule
\multirow{4}{*}{Systems} & Precision & BF16; FP32 for LLM head \\
& Tensor / pipeline / context parallelism & 2 / 1 / 4 \\
& GPU / Actor / rollout GPUs & H100 / 8 / 8 \\
& Bash tool timeout	& 600 s  \\
\midrule
\multirow{2}{*}{Evaluation} & Samples per task group & 4 \\
& Sampling temperature / top-$p$ & 0.7 / 0.95 \\
& Context / Completion maximum length & 128,000 / 20,480 tokens\\
\bottomrule[1.5pt]
\end{tabular}
\caption{Full RL training and evaluation hyperparameters, shared across all three methods unless explicitly distinguished.}
\label{tab:rl-hyperparameters}
\end{table*}

\subsection{Retrieval and Rubric Score Configuration}

For each agent trajectory, we form a query by combining the four views: task description, task stage, an LLM-extracted task contract, and an LLM-extracted state summary. We then retrieve the experience in the experience bank, where the experience is represented by both its full content and its LLM-extracted keywords using BM25 with $k_1=1.2$ and $b=0.75$. The eight rankings are fused by weighted reciprocal-rank fusion,
\begin{equation}
S_{\mathrm{ret}}(e)=\sum_{q\in\mathcal Q}\sum_{d\in\mathcal D}\frac{\alpha_q\beta_d}{60+\operatorname{rank}_{q,d}(e)}.
\label{eq:retrieval-score}
\end{equation}
The query weights for task description, stage, task contract, and state summary are $(0.60,0.69,1.48,4.01)$, respectively; the item weights for experience full content and keywords are $(0.19,1.39)$. We retrieve the top six experiences. The retrieval summarizer uses temperature 0.01 and a 4,096-token output limit.

Rubric scores are computed exactly as Eq. \ref{eq:rubric-score}. For the regression check $s_{x,r}(z(h))\geq s_{x,r}(p(h))$ in Algorithm~\ref{alg:cer-tts}, the judge compares each continuation directly with its parent prefix: a score of 0.5 means no progress over the parent, and a negative rubric contributes $(1-q_{ik})/2$, where $q_{ik}$ is its normalized score on continuation $h_i$, so that it penalizes only continuations that fail to beat this no-progress level. A continuation therefore regresses when its comparison score falls below 0.5. In RL training, we use 0.18 threshold to filter group without enough rubric score range. 

\section{Analysis (Cont')}

\subsection{More Metrics of TTS Experiments}
\label{app:more-tts-metrics}

\begin{table}[t]
\centering
\setlength{\tabcolsep}{6pt}
\renewcommand{\arraystretch}{1.15}
\begin{tabular}{@{}llcccc@{}}
\toprule[1.5pt]

&Step Budget & 20 & 40 & 60 & 80 \\
&Subset Size & 453/386 & 313/98 & 188/17 & 110/4 \\
\midrule[0.75pt]
\multirow{3}{*}{Nemotron 3 Ultra} & SWE-RM & 59.4\% & 52.4\% & 44.7\% & 40.9\% \\
& Adaptive Rubrics & 61.1\% & 54.0\% & 48.4\% & 43.6\% \\
& CER & \textbf{62.9\%} & \textbf{56.2\%} & \textbf{50.5\%} & \textbf{45.5\%} \\
\midrule
\multirow{3}{*}{Qwen 3.6 27B} & SWE-RM & 63.0\% & 44.9\% & - & - \\
& Adaptive Rubrics & \textbf{64.8\%} & 46.9\% & - & - \\
& CER & 63.7\% & \textbf{49.0\%} & - & - \\
\bottomrule[1.5pt]
\end{tabular}
\caption{RM@8 on fully unsubmitted subsets only. At each cutoff, we retain only tasks for which every candidate trajectory remains unsubmitted. All compared methods are evaluated on the same resulting task subset. We compare with the strongest two terminal reward baselines in the main TTS experiments, SWE-RM and Adaptive Rubrics. The two subset sizes are of model Nemotron 3 Ultra and Qwen 3.6 27B respectively. We report results only for subsets containing more than 50 tasks for statistical stability.}
\label{tab:tts-unfinished-rm}
\end{table}

In the results of Table~\ref{tab:tts-main}, at each step budget, the agent may already finish the task and submit a solution patch. Thus, the results at each step budget may not faithfully demonstrate CER's reward modeling capability on trajectory prefixes of various length. Table~\ref{tab:tts-unfinished-rm} reports the RM@8 on tasks that have all trajectory prefixes unsubmitted and compare with the two strong baselines on the same subset at each step budget. CER shows superior TTS performance at almost all trajectory length.

Table~\ref{tab:tts-same-repo-retrieval} reports the origin of retrieved experiences. Across different prefix budgets, 50.7–60.4\% of the retrieved experiences originate from the same repository as the target task. Thus, the current study primarily evaluates experience reuse within the SWE-bench domain; fully cross-repository experience transfer remains an open direction. We also observe that longer trajectories require more in-domain experience, since longer trajectories coming from the same parent prefix usually contains less variance and requires more discriminative and related experience.

Note that $\operatorname{Acc}_{\mathrm{pair}}$ is not directly comparable between CER and baseline methods since CER TTS expands only selected beam parents from the last round, which naturally leads to less variance within the group and more tied pairs. Moreover, the target value \(V_{x,\pi}(h_t)\) is latent: a single completed suffix provides only one Bernoulli sample rather than a ground-truth prefix value. Therefore, any pointwise classification accuracy will not be a reliable metric. This is the reason we only use RM@k to evaluate TTS performance. We also cannot report 95\% CI for TTS rollout-generation variance. Estimating this would require independently sampling multiple complete sets of eight long-horizon candidates for every task, which is computationally prohibitive in our setting.

\begin{wraptable}{r}{0.50\columnwidth}
\centering
\vspace{-10pt}
\scriptsize
\begin{tabular}{lrrrr}
\toprule[1.5pt]
Model & @20 & @40 & @60 & @80 \\
\midrule[0.75pt]
Nemotron 3 Ultra & 55.0\% & 55.7\% & 55.9\% & 58.8\% \\
Qwen 3.6 27B & 50.7\% & 51.4\% & 55.4\% & 60.4\% \\
\bottomrule[1.5pt]
\end{tabular}
\caption{The percent of retrieved experience summarized from the same-repository tasks in TTS experiments. CER leverages slightly more in-domain experience than out-of-domain experience.}
\label{tab:tts-same-repo-retrieval}
\vspace{-12pt}
\end{wraptable}

\subsection{TTS Components Ablation}
\label{sec:tts-ablation}

Table~\ref{tab:tts-ablation} ablates five components on both models, using the same sampled 100-instance SWE-bench Verified subset as Section \ref{sec:transfer}.

For the CER part, sibling comparison is the most important component. Removing it results in lowers average RM@8 by 3.0 pp on Qwen 3.6 27B and 4.0 pp on Nemotron 3 Ultra: this ablation retains the same candidate prefixes and retrieved experiences but generates and applies a separate rubric with only one continuation visible in each prompt, rather than jointly judging the sibling continuations sampled from the same parent. The drastic degradation at step budget 20 suggests that cross-continuation contrast is most useful when early candidates explore diverse hypotheses; descendants of the same selected beam become more behaviorally similar in later rounds, reducing the marginal benefit of joint judging. Removing retrieved experience lowers average RM@8 by 2.3 pp on Qwen 3.6 27B and 1.3 pp on Nemotron 3 Ultra, which indicates experience's benefit varies by how judge can effectively use it.

For the CER-guided search algorithm part, removing beam parents lowers average RM@8 by 2.3 pp on both models, and removing tie-breaking lowers it by 0.5 pp on Nemotron 3 Ultra and 0.8 pp on Qwen 3.6 27B. This indicates diversified samples are the most important component. Regression stopping lowers average RM@8 on Nemotron 3 Ultra by 1.5 pp when removed, but it does not help Qwen 3.6 27B, whose average improves by 1.3 pp without it. Most Qwen 3.6 27B trajectories submit within two or three rounds, so a pruned branch is rarely recovered by later search. Regression stopping is thus most useful for policies whose searches run longer, such as Nemotron 3 Ultra.

\subsection{Comparison between Memory-enhanced Evaluation and Generation}

\begin{wraptable}{r}{0.50\columnwidth}
\centering
\vspace{-15pt}
\scriptsize
\resizebox{\linewidth}{!}{\begin{tabular}{lrrrrr}
\toprule[1.5pt]
Method & @20 & @40 & @60 & @80 & Avg. \\
\midrule[0.75pt]
\multicolumn{6}{l}{\textit{Nemotron 3 Ultra}} \\
CER & \underline{63} & \textbf{68} & \textbf{68} & \textbf{69} & \textbf{67.0} \\
$-$ w/o tie-breaking & \underline{63} & \underline{67} & \textbf{68} & \underline{68} & \underline{66.5} \\
$-$ w/o regression & - & \underline{67} & \underline{66} & 66 & 65.5 \\
$-$ w/o beam parent & - & 66 & 65 & 65 & 64.8 \\
$-$ w/o experience & \textbf{65} & \underline{67} & 63 & \underline{68} & 65.8 \\
$-$ w/o sibling & 60 & 63 & 65 & 64 & 63.0 \\
\midrule[0.75pt]
\multicolumn{6}{l}{\textit{Qwen 3.6 27B}} \\
CER & \textbf{73} & \underline{75} & \underline{76} & \underline{75} & \underline{74.8} \\
$-$ w/o tie-breaking & \underline{72} & \underline{75} & 75 & 74 & 74.0 \\
$-$ w/o regression & - & \textbf{76} & \textbf{78} & \textbf{77} & \textbf{76.0} \\
$-$ w/o beam parent & - & 72 & 73 & 72 & 72.5 \\
$-$ w/o experience & 71 & 73 & 73 & 73 & 72.5 \\
$-$ w/o sibling & 65 & 74 & 74 & 74 & 71.8 \\
\bottomrule[1.5pt]
\end{tabular}}
\caption{TTS component ablation on a 100-instance SWE-bench Verified subset. Entries are resolved RM@8 (\%) at each budget $k$; regression and beam parents do not affect the first round (-), so Avg.\ uses the CER value at @20 for these rows. The best results per model are in \textbf{bold}, and the second-best results are \underline{underlined}.}
\label{tab:tts-ablation}
\vspace{-15pt}
\end{wraptable}

Although CER is designed for evaluation and is orthogonal to policy-side improvements, one may still ask whether memory-enhanced evaluation is necessary when the policy model already uses memory. Following SkillRL \citep{xia2026skillrl}, we implement an inference-time variant that distills successful and failed historical rollouts into a hierarchical skill bank and retrieves relevant general and task-specific skills during agent generation; we also use GPT-5.6 Sol to summarize these skills for a fair comparison. We cannot fully replicate the leave-one-out setup because SkillRL requires fusing skills from all historical tasks. We therefore exclude all 100 evaluated tasks during the fusion stage, ensuring that the policy never retrieves a skill derived from the instance it is solving.

Both memories are built from the same offline historical data. On the 100-task subset, the policy skill bank resolves 58\% of instances with Nemotron 3 Ultra and 68\% with Qwen 3.6 27B, slightly below average performance on the same subset (58.5\% and 68.1\%) without skill bank and even substantially below CER@20 (63\% and 73\%, respectively). A policy bank can only help a single generation, whereas an evaluation bank can leverage test-time search at a small additional cost: on Nemotron 3 Ultra, CER@20 uses only 26\% more tokens than a single rollout (3.96M versus 3.15M per instance; Table~\ref{tab:tts-main}). This gap reflects a fundamental asymmetry between generation and evaluation memory. SkillRL-style policy memory remains strongly task-dependent: the required skills vary by tasks, making it difficult to develop generalizable skills. Evaluation memory instead captures recurring behavioral evidence---such as efficient exploration, correct patch formatting, and not ignoring failed tests---that remains informative across repositories and tasks. The results show that even though CER shows out-of-domain generalization limitation in Table~\ref{tab:tts-same-repo-retrieval}, it is still much more generalizable than policy-focused memory.

\subsection{Training Dynamics}
\label{app:training-dynamics}

\begin{figure*}[t]
\centering
\includegraphics[width=0.96\textwidth]{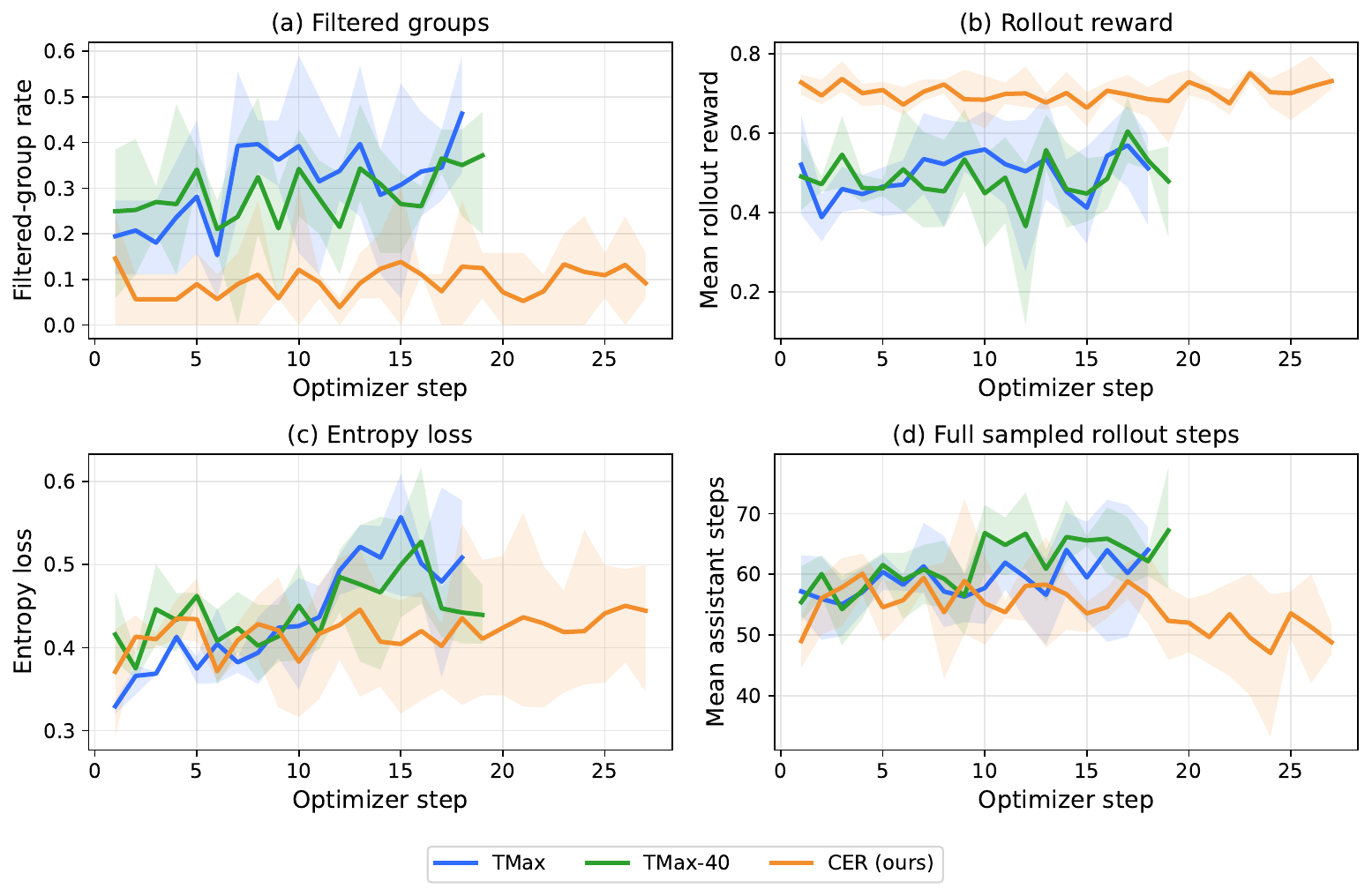}
\caption{Training dynamics of TMax, TMax-40, and CER. From left to right and top to bottom: the fraction of sampled groups filtered, mean reward over all sampled groups, entropy loss, and mean assistant steps in the complete sampled rollouts. Filtered-group rates include zero-variance groups but exclude discarded stale groups. The full-rollout step diagnostic does not represent the prefix length used for policy updates or actual training cost. We only rollout the prefix to terminal for analysis. Curves are means over three repository folds and shading spans the fold-wise minimum and maximum.}
\label{fig:training-dynamics}
\end{figure*}

Figure~\ref{fig:training-dynamics} reports the training dynamics of the exact runs used in Figure~\ref{fig:rl-token-efficiency}. Both TMax and TMax-40 filter substantially more groups than CER across the shared training horizon, showing that many terminal-verifier groups provide no relative learning signal. CER's entropy loss remains stable throughout training, while entropy loss under verifiable rewards rises markedly and becomes unstable in later steps. During training, CER shows a clear trend toward shorter rollouts, while TMax and TMax-40 show the opposite trend. This may suggest that rubrics encourage more efficient and cleaner behavior. Because TMax and TMax-40 use verifier rewards whereas CER uses rubric rewards, their absolute reward levels are not directly comparable; the relevant evidence is each curve's temporal stability together with the main downstream results. The mean rollout reward stays flat for all methods, as expected: each optimizer step samples different tasks and dynamic sampling changes the batch composition, so this curve reflects batch composition rather than learning progress, which is tracked by the validation curves in Figure~\ref{fig:rl-token-efficiency}.

\paragraph{Denser signal.} Through the matched 384-step checkpoint, 43.8\% of CER's sampled groups have identical binary resolution across all eight full trajectories, so binary terminal outcomes provide no relative signal for them. CER still trains on 83.3\% of these groups because their rubric scores differ (mean weighted rubric range 0.43, against the abstention threshold of 0.18); they account for 40.5\% of all groups CER trains on.

\paragraph{Reward hacking.} Shorter CER rollouts do not come from premature submission. During training, CER submits within the first 40 turns less often than TMax (27.1\% versus 32.7\% of sampled rollouts), and this rate does not increase through the matched 384-step checkpoint (24.8\% in the first half versus 23.9\% in the second). Using regular-expression proxies over the first 40 turns, success claims made without executing any program are rare and equally so for all methods (0.4--0.6\%).

\subsection{Can CER be Used as Process Reward Model?}

A process reward must be comparable across different points of the same trajectory, whereas CER is designed to rank prefixes observed at a shared cutoff. We sample one random Nemotron 3 Ultra rollout for each of the 500 SWE-bench Verified instances (500 rollouts in total) and score each prefix every 10 steps through step 100 without group-wise comparison. For each instance, we fix the six experiences most frequently retrieved in the main TTS experiment and reuse this same set at every cutoff, so changes across steps cannot be attributed to retrieval drift. We treat the normalized rubric score as the predicted probability of terminal resolution and measure binary cross-entropy against the terminal label. On this matched cohort, the mean score rises from 0.431 at step 10 to 0.852 at step 100 even though the terminal labels are unchanged, while binary cross-entropy increases from 0.755 to 3.514. The rubric judge therefore becomes progressively more confident without becoming correspondingly more predictive of terminal success. Step-specific grouping or calibration could remove this shift, but would require a group sampling at each step. In its current form, CER is therefore suitable as a value model for comparing candidates at a fixed prefix length, but not as a general process reward model.

\subsection{How Early Can We Predict?}
\label{sec:turn-ablation}

To know how early rubric reward can be comparable to verifier reward, we train otherwise identical CER policies on fold 0 with step budgets of 10, 20, 30 and 40 steps (we reuse the main experiments results for 40 step). Their test set resolved rates are 48.1\%, 48.3\%, 49.8\%, and 52.7\%, respectively. To interpret these budgets, we annotate one randomly sampled Qwen 3.5 9B rollout for each of the 500 SWE-bench Verified instances (500 rollouts in total). The median entry steps for editing and post-edit validation are 22 and 28, respectively. Thus, 10 / 20 steps often expose only exploration or the beginning of editing, whereas at 30 / 40 steps the trajectory can include both implementation and validation evidence, consistent with their stronger downstream results.

We also find that early exploration behavior alone is not a good indicator of final success. To show this, we compute at every step through 100 the fraction of reference-patch target files touched so far and use this fraction directly to predict terminal resolution on the same samples. The AUROC peaks at only 0.608 at step 25 as shown in Appendix Figure~\ref{fig:early-prediction-prm}(a). Even with access to the ground-truth patch, the files touched during exploration carry limited outcome information by themselves: failed trajectories can inspect the right files without producing a correct implementation. Final reward therefore cannot be predicted arbitrarily early: prefixes that include editing and validation evidence are necessary.

\begin{figure*}[t]
\centering
\includegraphics[width=\textwidth]{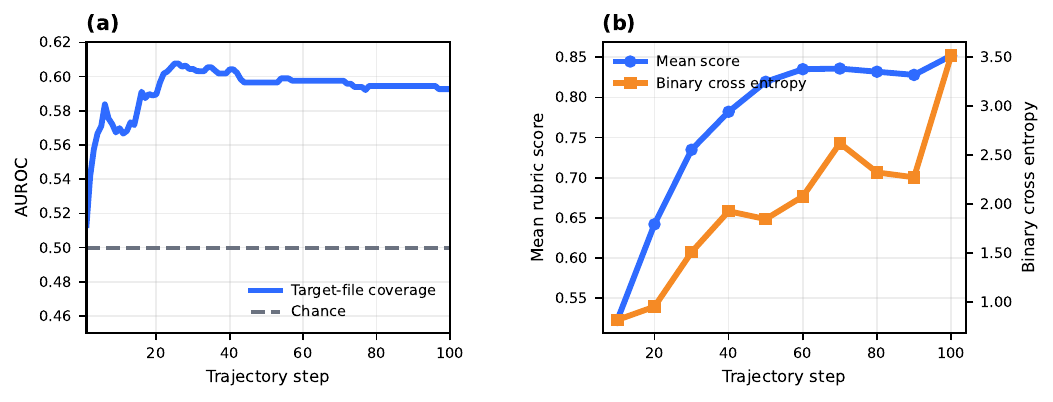}
\caption{(a) AUROC of using the cumulative fraction of golden-patch target files touched to predict terminal verifier reward, computed on one sampled Qwen 3.5 9B rollout for each of the 500 SWE-bench Verified instances. The cumulative fraction of golden-patch target files touched is a weak indicator of terminal reward and can be predictive only after step 20. (b) Mean normalized rubric score and binary cross-entropy against terminal verifier reward, computed on one sampled Nemotron 3 Ultra rollout for each of the 500 SWE-bench Verified instances. Lower binary cross-entropy is better. Nemotron judged scores exhibit strong length bias: longer trajectories receive higher scores which lead to high inaccuracy as the trajectory length increase.}
\label{fig:early-prediction-prm}
\end{figure*}

\begin{table*}[t]
\centering
\footnotesize
\setlength{\tabcolsep}{4pt}
\renewcommand{\arraystretch}{1.3}
\begin{tabular}{@{}p{0.25\textwidth}>{\centering\arraybackslash}p{0.2\textwidth}p{0.5\textwidth}@{}}
\toprule[1.5pt]
\textbf{Primary category} & \textbf{Weighted percent} & \textbf{Example} \\
\midrule[0.75pt]
Pre-edit reproduction and runnable testbed & 0.65\% & For Django-13033, the criterion requires executed evidence that exposes the erroneous SQL join and ordering direction and contrasts it with semantically informative control orderings. \\
Root-cause localization and causal diagnosis & 16.82\% & For Astropy-12907, the criterion requires model to identify that recursive separability yields an array for the nested right operand and that \texttt{\_cstack} destroys its dependency structure by filling the embedded block with ones. \\
Fix-design convergence and authoritative grounding & 0.36\% & For scikit-learn-9288, the criterion goes beyond locating the serial--parallel seed mismatch by requiring a branch-invariant repair that draws one deterministic integer seed per \texttt{n\_init} run and passes the corresponding seed to every serial invocation. \\
Investigation efficiency and environment recovery & 3.43\% & For pytest-7432, the criterion requires recovery from unusable or contradictory runtime evidence by identifying which pytest installation is executing, avoiding contamination of installed packages, and returning to authoritative repository evidence without overstating validation. \\
Intermediate-edit safety and workspace hygiene & 1.29\% & For Django-10914, the criterion preserves independent directory defaults and secure temporary-file behavior while applying mode 0644 only to final stored uploads.\\
\bottomrule[1.5pt]
\end{tabular}
\caption{Semantic taxonomy of rubric criteria, Part I: exploration and diagnosis. Percentages are shares of the total rubric weight over all rubrics. Each example summarizes a real criterion together with its applicability.}
\label{tab:rubric-taxonomy-exploration}
\end{table*}

\begin{table*}[t]
\centering
\footnotesize
\setlength{\tabcolsep}{4pt}
\renewcommand{\arraystretch}{1.3}
\begin{tabular}{@{}p{0.25\textwidth}>{\centering\arraybackslash}p{0.2\textwidth}p{0.5\textwidth}@{}}
\toprule[1.5pt]
\textbf{Primary category} & \textbf{Weighted percent} & \textbf{Example} \\
\midrule[0.75pt]
Functional implementation correctness & 18.07\% & For Astropy-13398, the criterion requires a coherent implementation of bidirectional ITRS--AltAz and ITRS--HADec transforms using topocentric geometry while keeping the terrestrial position time-invariant. \\
Edge-case and implementation completeness & 5.81\% & For Django-13089, the criterion covers the empty-result edge case in which a culling query returns no row: execution must complete without subscripting \texttt{None} or deleting a valid cache key. \\
Patch scope and minimality & 5.47\% & For Astropy-12907, the criterion confines the repair to \texttt{\_cstack}: copy the computed right separability matrix into its positioned block rather than overwrite that block with ones.  \\
Compatibility and integration preservation & 8.88\% & For Astropy-14309, the criterion requires the FITS identifier to return false when no positional object exists without indexing an empty tuple, while preserving signature, extension, and HDU detection. \\
Final-code quality and maintainability & 0.47\% & For SymPy-16450, the criterion requires \texttt{posify} to transfer compatible known assumptions through replacement-symbol construction, give \texttt{positive=True} unambiguous precedence, and avoid private-state mutation.  \\
Submission and patch-artifact integrity & 9.48\% & For Django-14915, the criterion requires a complete, repository-targeted patch in a format that the evaluator can directly recognize and ingest, rather than prose or partial diff fragments that require reconstruction. \\
\bottomrule[1.5pt]
\end{tabular}
\caption{Semantic taxonomy of rubric criteria, Part II: implementation and editing. Percentages and examples follow Table~\ref{tab:rubric-taxonomy-exploration}.}
\label{tab:rubric-taxonomy-implementation}
\end{table*}

\begin{table*}[t]
\centering
\footnotesize
\setlength{\tabcolsep}{4pt}
\renewcommand{\arraystretch}{1.3}
\begin{tabular}{@{}p{0.25\textwidth}>{\centering\arraybackslash}p{0.2\textwidth}p{0.5\textwidth}@{}}
\toprule[1.5pt]
\textbf{Primary category} & \textbf{Weighted percent} & \textbf{Example} \\
\midrule[0.75pt]
Causal before/after verification & 1.17\% & For Django-11206, the criterion requires executed post-edit evidence for the reported tiny-Decimal behavior plus counterexamples that expose raw-exponent zeroing, lost scientific safeguards, or bypassed formatting arguments. \\
Targeted behavioral validation & 14.43\% & For Astropy-13236, the criterion requires evidence that direct multi-field array input reaches ordinary \texttt{Column} construction without automatic \texttt{NdarrayMixin} conversion while explicit columns and legitimate mixins retain established handling.\\
Edge-case and scenario validation & 0.88\% & For Django-13028, the criterion distinguishes an ordinary lookup RHS whose application attribute is false from an ORM expression that intentionally prohibits filtering, using source tracing or a task-shaped boundary experiment. \\
Regression-suite and breadth validation & 11.64\% & For Astropy-14096, the criterion must establish the corrected nested-property error while protecting ordinary missing attributes and SkyCoord's dynamic frame and transform behavior. \\
Execution-environment fidelity & 0.57\% & For Django-11880, the criterion credits the dictionary-isolation check only when it is executed against the edited checkout, rather than a toy object or a different installed version. \\
Failure interpretation and claim discipline & 0.34\% & For Django-12193, the criterion requires a consistent ledger in which a failed property is resolved only after its implementation or oracle is corrected and rerun, and success claims remain limited to demonstrated properties. \\
Security, performance, and other nonfunctional risk & 0.25\% & For Django-13195, the criterion requires cookie expiry to emit the intended SameSite value while preserving expiration, caller-selected path and domain, protected-name security, and generic deletion behavior. \\
\bottomrule[1.5pt]
\end{tabular}
\caption{Semantic taxonomy of rubric criteria, Part III: validation and other nonfunctional risks. Percentages and examples follow Table~\ref{tab:rubric-taxonomy-exploration}.}
\label{tab:rubric-taxonomy-validation}
\end{table*}

\clearpage
\newpage

\section{Prompt Templates}
The following cards reproduce the complete prompts used by CER.

\begin{promptcard}{Trajectory Summary}
You are maintaining a compact, durable working memory for a long-running software-debugging trajectory.

## Goal
Update the persistent state after older trajectory segments are evicted. Preserve the most important actionable context needed for later judging, and continuation of the work.

## Required Sections
Return exactly these 8 top-level string fields:
- **current_state**: What is actively being worked on right now, pending tasks, and immediate next steps. Always refresh this section so it reflects the latest work.
- **task_specification**: What the user asked for, important constraints, acceptance criteria, design decisions, and explanatory context.
- **files_and_functions**: Important files, functions, classes, modules, and why they matter. Include concrete file paths and identifiers.
- **errors_and_corrections**: Errors encountered, failed attempts, rejected hypotheses, and how they were corrected. Record approaches that should not be retried.
- **codebase_and_system_documentation**: Important components, interfaces, workflows, or architectural relationships and how they fit together.
- **learnings**: Actionable lessons about what worked well, what did not, and what to avoid. Do not duplicate material already captured in other sections.
- **key_results**: Exact or near-exact outputs that should be preserved, such as a patch idea, a concrete answer, a command result, or another critical artifact.
- **worklog**: Very terse step-by-step record of what was attempted or completed.

## Writing Guidelines
- Keep only information supported by the previous state, the evicted trajectory, or the workspace metadata.
- Be detailed and information-dense. Include concrete file paths, function names, commands, test names, error messages, patch details, and technical observations when useful.
- Focus on actionable, specific context that would help someone understand, judge, or recreate the work.
- It is OK to leave a section unchanged or blank if there are no substantial new insights. Do not add filler such as "No info yet".
- Keep each section under 400 words. If a section gets too long, remove lower-value details while preserving the most decision-relevant information.
- Preserve older facts that still matter.
- Merge redundant details instead of repeating them.
- If an earlier belief was revised, record that correction explicitly in the appropriate section.
- Do not hallucinate. Prefer omission to speculation.

## Output Format Example
<format_example>

THOUGHT: <your reasoning process>

```json
{
  "current_state": "",
  "task_specification": "",
  "files_and_functions": "",
  "errors_and_corrections": "",
  "codebase_and_system_documentation": "",
  "learnings": "",
  "key_results": "",
  "worklog": ""
}
```

</format_example>

## Inputs
1. **Question**: Original system and user prompt containing the coding task
2. **Previous Persistent State**: Previous memory state
3. **Evicted Older Trajectory**: Older trajectory segments that must now be compressed
4. **Workspace Metadata**: Compact git-based metadata at the current step

## Question:
 System Prompt:
{system_prompt}
 User Prompt:
{user_prompt}

## Previous Persistent State:
{previous_persistent_state}

## Evicted Older Trajectory:
{evicted_older_trajectory}

## Workspace Metadata:
{workspace_metadata}
\end{promptcard}

\begin{promptcard}{Rubric Generation}
You are an expert evaluator generating adaptive rubrics to assess agent trajectory continuations.

## Task
Identify the single most discriminative rubric to output next for judging the current trajectory continuations. Capture subtle quality differences that existing rubrics miss.
This is a multi-turn rubric generation setting. At each turn, output exactly one rubric object or an empty JSON object `{}`. The rubric object may be either a newly generated rubric or a reused/adapted existing rubric. Existing Rubrics contains previously generated rubrics that may be reused/adapted and should be used to understand the current evaluation gap and avoid redundancy.
Every generation sample must output at least one non-empty rubric before it may return `{}`. Existing rubrics are not reused automatically; reuse requires outputting the full rubric object again.
If no further rubric should be output in this multi-turn generation process, return an empty JSON object: {}.

## Output Components
- **Title**: Concise abstract label (general, not task-specific)
- **Description**: Detailed, specific description of what makes a continuation excellent/problematic
- **Scale**: A five-point scale from 1 to 5 with concrete anchors for this rubric. The scale must follow the rubric polarity: for a positive rubric, 1 is the weakest evidence and 5 is the strongest evidence; for a negative rubric, 1 is no/least evidence of the flaw and 5 is the most severe evidence of the flaw.
- **Polarity**: Either `"positive"` or `"negative"`.
- **Weight**: A positive number expressing this rubric's relative importance among all the rubrics generated. Use `1.0` as a neutral default and assign higher weights to impactful rubrics and lower weights to less impactful ones. Weight is always positive even for negative rubrics.
- **Metadata**: A structured evidence payload for style-specific extra content. Use string fields such as `stage`, `oracle_test`, `code_review`, `privileged_reference_summary`, `judge_focus`, or `failure_mode`. Put complete test snippets, review reasoning, or reference-derived behavioral oracles here instead of overloading the title or scale.

## Categories
A rubric may be either:
1. **Positive Rubrics**: Excellence indicators distinguishing superior continuations
2. **Negative Rubrics**: Critical flaws definitively degrading quality
Represent this choice using the `polarity` field in the rubric object.

## Core Guidelines

### 1. Discriminative Power
- Focus ONLY on criteria meaningfully separating quality levels
- Each rubric must distinguish between otherwise similar continuations from the same shared prefix
- Exclude generic criteria applying equally to all continuations

### 2. Grounded Current Distinction
- Focus the rubric on the differences between the continuations, not on the shared context
- If continuations differ in diagnostic strategy, reproduction attempts, validation attempts, or targeting of relevant files, define a rubric around that concrete process evidence
- If continuations differ in source changes, define a rubric around the observable patch behavior: changed files, symbols, API contracts, data flow, compatibility boundaries, edge cases, or tests
- If continuations share the same visible core behavior, do not separate them using harmless formatting, error-message wording, local variable placement, scratch scripts, or transient test scaffolding unless those details create an observable behavioral risk
- Do not create standalone style rubrics for DRYness, helper extraction, formatting, comments, or cleanup. Such details are only valid when the visible diff shows a concrete behavioral, compatibility, or maintainability risk that affects the task outcome
- Do not reward majority behavior just because most continuations share it; reward the behavior best supported by the visible evidence
- Avoid vague criteria such as "thoroughness", "correctness", "best practice", or "complete implementation" unless the rubric defines the concrete evidence being scored

### 3. Novelty & Non-Redundancy
- Do not generate a new rubric that duplicates any existing or already generated rubric in meaning/scope. Re-outputing an existing rubric as a reused/adapted rubric is allowed and is not considered duplication.
- Identify uncovered quality dimensions
- Add granular criteria if existing rubrics are broad
- Return `{}` only when no remaining existing rubric should be reused/adapted and no new non-redundant rubric should be generated
- Do not generate semantically equivalent rubrics, e.g., "Runs targeted validation" as a positive rubric and "Does not run targeted validation" as a negative rubric.
- Choose only the more discriminative direction

### 4. Conservative Negative Rubrics
- Identify clear failure modes, not absence of excellence
- A negative rubric should describe an observable harmful behavior, incorrect assumption, misleading edit, or unsupported claim
- Do not create a negative rubric merely because a continuation lacks a desirable behavior
- For negative rubrics, every scale anchor must measure severity of the flaw: 1 means the flaw is absent or minimal, and 5 means the flaw is clearly and severely present. Never write a negative rubric whose scale rewards the good behavior at 5.

### 5. Previous Generated Rubrics & Experiences
- In each turn, output exactly one rubric object or `{}`. A non-empty output may either reuse/adapt a previously generated rubric if it is still applicable and discriminative for the current trajectory continuations, or generate one new rubric, optionally informed by retrieved experiences.
- If a concrete, important trajectory behavior gap is not captured by any existing rubric or experience, generate a new rubric for that missing evaluation criterion. This is especially needed when you identify task-specific mistakes at the specific agent stage.
- Rubric reuse is opt-in, not automatic. Previously generated rubrics are not used by the judge merely because they appear under Existing Rubrics. To reuse an existing rubric, output the full rubric object in one generation turn. Any existing rubric that is not output in the current multi-turn generation process is considered dropped and will not participate in judging.
- You may freely modify `weight` or other details of an existing rubric or reference golden rubric in the experience to better reflect the judging importance and focus at the current stage. A modified existing rubric still counts as reuse/adaptation as long as it is derived from an existing rubric.
- Use previous rubrics and experiences to understand what is already covered, then output only one useful next rubric: either a reused/adapted existing rubric or a non-redundant uncovered criterion.
- An experience is guidance about when a rubric is useful or misleading. Treat its `context` and `experience` fields as applicability conditions, not as facts about the current task.
- `metadata.reference_golden_rubrics` contains candidate criteria learned from earlier cases. You may adapt a candidate's polarity, scope, wording, metadata, scale, or weight only when the current continuations match the prior lesson.
- Do not copy a reference rubric merely because it was retrieved. A copied or adapted rubric must be independently relevant and judgeable from the current task and trajectory continuations.
- Retrieved experiences are not privileged oracles. Never infer the current final patch, hidden tests, or outcome from their presence.
- If a previous rubric or experience would turn a precise current distinction into a broad generic or misleading rubric, ignore it.

## Selection Strategy

### Quantity: in each turn, output exactly one rubric object or return `{}`. Across the full multi-turn generation process, output at least 1 and at most 6 non-empty rubrics in total.
- Output exactly one rubric object if the next rubric should participate in judging, whether it is newly generated or reused/adapted from existing rubrics.
- Return `{}` only when no remaining existing rubric should be reused/adapted and no new high-impact, non-redundant rubric should be generated.

### Polarity Selection Based on Response Patterns:
- **More positive**: When continuations lack sophistication but avoid major errors
- **More negative**: When systematic failure patterns are present
- **Balanced across turns**: When both excellence gaps and failure modes exist
- **Empty object**: When no further rubric should be output: all useful existing rubrics have already been reused/adapted in previous generation turns, and no new high-impact, non-redundant rubric remains

## Analysis Process
1. Group continuations by quality level
2. Find factors separating higher/lower clusters
3. Check if factors are covered by rubrics already output in the current multi-turn generation process
4. Select the single criterion with the highest discriminative value

## Output Format Example
<format_example>

THOUGHT: <your reasoning process>

```json
{
    "polarity": "<positive|negative>",
    "weight": <positive number>,
    "description": "<detailed excellence/failure description>",
    "title": "<abstract label>",
    "metadata": {
      <a dict of any other relevant structured context and evidence needed for judging, including but not limited to current working stage, focus, targeted test code or pseudo-test, code review, etc.>
    },
    "scale": {
      "1": "<positive rubric: weakest evidence / negative rubric: no evidence of the flaw>",
      "2": "<positive rubric: weak evidence / negative rubric: minor evidence of the flaw>",
      "3": "<the moderate anchor>",
      "4": "<positive rubric: strong evidence / negative rubric: clear evidence of the flaw>",
      "5": "<positive rubric: strongest evidence / negative rubric: most severe evidence of the flaw>"
    }
}
```

</format_example>

If no further rubric should be output in this multi-turn generation process, output:
<format_example>

THOUGHT: <your reasoning process>

```json
{}
```

</format_example>

## Inputs
1. **Question**: Original system and user prompt containing code problem statement
2. **Previous Persistent State**: Current memory state with summary of past findings and milestones
3. **Parent Trajectory**: The most recent agent trajectory
4. **Agent Trajectory Continuations**: Multiple agent trajectories continued from the latest trajectory (Continuation 1, Continuation 2, etc.)
5. **Existing Rubrics** (optional): Previously generated rubrics available for reuse/adaptation and redundancy checking. They are not automatically used for judging unless explicitly output as full rubric objects in the current multi-turn generation process.

## Critical Reminders
- Each rubric must distinguish between the actual provided continuations
- Exclude rubrics applying equally to all continuations
- Prefer `{}` over redundant rubric generation when no remaining existing rubric should be reused/adapted and no new non-redundant rubric remains
- Focus on observable, objective, actionable criteria
- Quality over quantity: 1 excellent rubric > multiple mediocre ones
- The shared context is common to all continuations. Focus the rubric on differences between the continuations themselves
- Do not return `{}` when there is still a useful existing rubric to reuse/adapt or a visible, important, non-redundant difference in diagnostic strategy, reproduction attempts, validation attempts, or targeting of relevant files
- Never output a list of rubrics. Each generation turn must output exactly one rubric object or `{}`
- Output in the required format. Do not restate the question, previous state, agent trajectories, or existing rubrics in the response.

Generate only the most impactful, non-redundant rubric revealing meaningful quality differences, or explicitly reuse/adapt one existing rubric that should participate in judging.

## Question:
System Prompt:
{system_prompt}

User Prompt:
{user_prompt}

## Previous Persistent State:
{previous_persistent_state}

## Parent Trajectory:
{parent_trajectory}

## Agent Trajectory Continuations:
## Continuation {i}:
{continuation_i}

## Existing Rubrics:
{existing_rubrics}

## Retrieved Rubric Experiences:
{retrieved_experiences}

[FOLLOW-UP USER MESSAGE AFTER EACH NONEMPTY RUBRIC]
Output the next rubric that should participate in judging, either newly generated or reused/adapted from existing rubrics, or return an empty object `{}` to stop. Output exactly one rubric object or `{}`; never output a list.
\end{promptcard}

\begin{promptcard}{Rubric Judging}
You are an expert evaluator scoring one agent trajectory continuation given one rubric.

## Task
Evaluate the provided continuation trajectory using the provided criterion and the shared context.

## Core Guidelines
- Judge only the specified criterion, not general quality
- Use the rubric's scale exactly as being required. For negative rubrics, the scale is inverted (e.g. worst case should receive 5 while best case should receive 1)
- Score the continuation trajectory itself, not the underlying task or bug in the abstract
- Use only evidence visible in the continuation trajectory. Do not hallucinate or infer unstated facts
- Use the previous persistent state and latest agent trajectory only when it is needed to interpret the continuation
- Ground the score in exact visible trajectory evidence. In the final JSON object, put the evidence field before the score field.
- Keep the structured answer limited to the evidence and score fields. Do not restate the full question, criterion, persistent state, or trajectories inside the JSON block.

## Output Format Example
<format_example>

THOUGHT: <your reasoning process>

```json
{
  "evidence": "<exact visible evidence supporting the selected scale anchor>",
  "score": <a score on a scale of 1 to 5 indicating how appropriate the continuation is based on the scale of the given criterion>
}
```

</format_example>

## Inputs
1. **Question**: Original system and user prompt containing code problem statement
2. **Previous Persistent State**: Current memory state with summary of past findings and milestones
3. **Parent Trajectory**: The most recent agent trajectory
4. **Continuation Trajectory**: A agent trajectory continued from the latest trajectory
5. **Criterion**: The specific criterion to evaluate

## Question:
System Prompt:
{system_prompt}

User Prompt:
{user_prompt}

## Previous Persistent State:
{previous_persistent_state}

## Parent Trajectory:
{parent_trajectory}

## Continuation Trajectory:
{continuation_trajectory}

## Criterion:
{criterion}
\end{promptcard}

\begin{promptcard}{Experience Generation}
[SYSTEM MESSAGE]
You diagnose one concrete SWE trajectory-selection failure and write one task-specific experience for a second-pass evaluator. Reason freely, then append exactly one JSON object matching the requested schema.

[USER MESSAGE TEMPLATE]
## Objective
Create one deliberately narrow experience for a second-pass rubric portfolio editor after initial rubrics are generated and before any replacement scores are produced. It must correct this round's observed coverage, semantic alignment, redundancy, weight allocation, stage alignment, tie discipline, or displacement of the primary contract failure and is evaluated alone in the `{scope}` bank.

The privileged diagnostic packet may be used only to discover the bias. The public card must state what visible evidence to inspect and how to act. It must not mention rewards, hidden tests, node IDs, instance IDs, golden patches, this experiment, or future outcomes. Do not prescribe an exact patch when equivalent implementations are valid. Include:

- a narrow activation condition;
- decisive positive and contradictory evidence;
- close-looking cases that must be excluded;
- an explicit abstention rule when the prefix cannot distinguish candidates.

Do not merely restate the issue. End with one JSON object, either directly or under
`content`, matching:

{
  "title": "short reusable title",
  "description": "when this experience should be considered",
  "context": "visible activation, exclusion, and abstention boundaries",
  "experience": "specific evidence-reading or rubric-generation policy",
  "metadata": {"reference_golden_rubrics": []}
}

## Attempt
{attempt} of 2

## Privileged diagnostic packet
{privileged_diagnostic_packet}

## Visible paused judge view
{visible_paused_judge_view}
\end{promptcard}

\begin{promptcard}{Experience Refinement}
[SYSTEM MESSAGE]
You are the refinement stage for SWE trajectory-judge
experiences. Correct the measured failure of the previous card without leaking
privileged outcomes. Reason freely, then append exactly one valid JSON object.

[USER MESSAGE TEMPLATE]
## Objective
Create one deliberately narrow experience for a second-pass rubric portfolio editor after initial rubrics are generated and before any replacement scores are produced. It must correct this round's
observed coverage, semantic alignment, redundancy, weight allocation, stage alignment, tie discipline, or displacement of the primary contract failure and is evaluated alone in the `{scope}` bank.

The privileged diagnostic packet may be used only to discover the bias. The public
card must state what visible evidence to inspect and how to act. It must not mention
rewards, hidden tests, node IDs, instance IDs, golden patches, this experiment, or
future outcomes. Do not prescribe an exact patch when equivalent implementations are
valid. Include:

- a narrow activation condition;
- decisive positive and contradictory evidence;
- close-looking cases that must be excluded;
- an explicit abstention rule when the prefix cannot distinguish candidates.

Do not merely restate the issue. End with one JSON object, either directly or under
`content`, matching:

{
  "title": "short reusable title",
  "description": "when this experience should be considered",
  "context": "visible activation, exclusion, and abstention boundaries",
  "experience": "specific evidence-reading or rubric-generation policy",
  "metadata": {"reference_golden_rubrics": []}
}

## Attempt
{attempt} of 2

## Privileged diagnostic packet
{privileged_diagnostic_packet}

## Visible paused judge view
{visible_paused_judge_view}

## Previous attempted card and measured result
{previous_attempt_and_measured_result}
It failed to improve both target metrics. Diagnose why it did not alter the rubric/judge behavior and make a substantive correction, not a paraphrase.
\end{promptcard}

\begin{promptcard}{RL Direct Abstention}
[SYSTEM MESSAGE]
You are the conservative same-reward gate for an early-trajectory reward predictor. The eight continuations are independent prefixes from the same root. Judge only their current visible behavior, command observations, repository state, tests, evaluator-facing artifacts, and the supplied golden reference rubrics. Never predict later actions or outcomes.

`should_abstain=true` requires strong positive visible evidence that all eight prefixes are materially equivalent for final evaluator reward. If any prefix has a credible task-relevant semantic, artifact, failure, recovery, or discriminative-validation difference, return `should_abstain=false`. If the supplied reference rubrics do not cover a visible difference, do not assume equality; return false. Optimize for recall of truly different-reward groups because a false abstention cannot be recovered by the numeric detector.

Workflow stage, exploration depth, verbosity, generic validation volume, and harmless temporary files do not establish reward equality or difference by themselves. A reversible intermediate mistake is not a terminal flaw after visible recovery, and an unfinished prefix is not an empty submission.

For `terminal_at_cutoff=true`, the terminal assistant payload is the
evaluator-facing submission. A `non_patch` terminal payload cannot be rescued
by a valid workspace source or prose. If one terminal prefix is `unified_diff`
and another is `non_patch`, that visible format difference must block
abstention. `terminal_at_cutoff=false` means terminal-artifact criteria are
inapplicable.

Return exactly one JSON object:
{
  "should_abstain": <boolean>,
  "confidence": <number from 0 to 1>,
  "same_reward_evidence": ["visible evidence supporting material equivalence"],
  "variance_evidence": ["visible evidence that prevents abstention"]
}

[USER MESSAGE TEMPLATE]
## Task
{task}

## Golden reference rubrics
{"reference_golden_rubrics": {golden_reference_rubrics}}

## Prefix group
{prefix_group}
\end{promptcard}

\begin{promptcard}{RL Teacher Rubric Generation}
You create the initial instance-specific golden rubric portfolio for SWE early-trajectory judging. You receive the task and every frozen current-batch rollout-40 group for this instance. Each group contains cutoff-visible trajectories and private final joint values used only as supervision.

Return one to six complete, complementary reference golden rubrics. Use the available capacity when the task has several independently observable causal dimensions; do not under-specify a complex task. Useful dimensions commonly include task-causal diagnosis, the precise implementation boundary, decisive tests or runtime evidence, compatibility/regression protection, and recovery from contradictory evidence. Do not create redundant rubrics or split one dimension merely to fill the limit.

Each rubric must independently help order the supplied trajectories and remain valid for future rollouts of this same task. Focus on observable evidence at the cutoff. Never encode aliases, group IDs, private values, reward ordering, future actions, final outcomes, model names, or stage position as a proxy. Legitimate ties remain ties. Do not claim that proposed work was applied or tested.

The individual judge sees title, polarity, description, metadata, and scale. Put precise evidence sources, precedence, prerequisites, non-evidence, and contradiction caps in metadata and mutually exclusive scale anchors. `applies_when` is visible only to the rubric generator and must contain task or evidence selection conditions, never instructions to judge, inspect, score, weight, grade, compare, or verify evidence.

Every rubric has this complete schema:
{
  "title": "...",
  "polarity": "positive or negative",
  "weight": 1.0,
  "applies_when": ["selection condition only"],
  "description": "observable criterion",
  "metadata": {
    "stage": "...",
    "judge_focus": "specific visible evidence",
    "evidence_authority": "which visible evidence wins conflicts",
    "hard_gate": "prerequisite for higher anchors",
    "contradiction_rule": "visible evidence that caps credit",
    "oracle_test": "task-specific semantic reference",
    "failure_mode": "common misjudgment this rubric prevents"
  },
  "scale": {"1":"...","2":"...","3":"...","4":"...","5":"..."}
}

Return exactly one JSON object with exactly one top-level field:
{"reference_golden_rubrics": [rubric, ...]}
Return no handbook, stages, behavior lists, abstain guidance, weights policy, analysis, or prose outside that JSON object.

## Instance packet
{
  "task": {task},
  "historical_rollout_groups": {historical_rollout_groups}
}
\end{promptcard}

\begin{promptcard}{RL Teacher Rubric Refinement}
You repair failed candidate golden rubrics for one fixed SWE instance. Each candidate was already judged by judge model on every current-batch rollout-40 group. You receive its full wrong-pair evidence, private final values used only as supervision, the complete current portfolio, and visible trajectory views.

For every supplied failed candidate, return exactly one corrected rubric. Deletion and abstention are forbidden at this stage: portfolio pruning belongs only to downstream weight optimization through zero weight and the six-rubric limit. Preserve the candidate's intended semantic dimension. Use judging evidence to repair evidence authority, hard gates, contradiction caps, mutually exclusive anchors, polarity, or scope. Even when the repair may fail, make the best truthful cutoff-observable attempt.

Never encode aliases, group IDs, judge scores, private values, reward ordering, future actions, final outcomes, model names, or progress proxies. `applies_when` contains only rubric-generator selection conditions.

Output exactly one JSON object:
{
  "decisions": [
    {
      "candidate_id": "exact supplied id",
      "action": "refine",
      "reason": "specific causal diagnosis and attempted repair",
      "rubric": {
        "title": "...",
        "polarity": "positive or negative",
        "weight": 1.0,
        "applies_when": ["selection condition only"],
        "description": "observable criterion",
        "metadata": {
          "stage": "...",
          "judge_focus": "...",
          "evidence_authority": "...",
          "hard_gate": "...",
          "contradiction_rule": "...",
          "oracle_test": "...",
          "failure_mode": "..."
        },
        "scale": {"1":"...","2":"...","3":"...","4":"...","5":"..."}
      }
    }
  ]
}
Return every candidate_id exactly once and no prose outside JSON.

## Instance packet
{
  "task": {task},
  "current_reference_golden_rubrics": {current_reference_golden_rubrics},
  "current_rollout_groups": {current_rollout_groups}
}

## Failed candidate judge outcomes
{failed_candidate_judge_outcomes}
\end{promptcard}

\end{document}